\documentclass{article}

\usepackage{arxiv}

\usepackage[utf8]{inputenc} 
\usepackage[T1]{fontenc}    
\usepackage{hyperref}       
\usepackage{url}            
\usepackage{booktabs}       
\usepackage{amsfonts}       
\usepackage{nicefrac}       

\usepackage{graphicx}
\usepackage{natbib}
\usepackage{doi}
\usepackage{amsmath}
\usepackage{amssymb}
\usepackage{subcaption} 
\usepackage{adjustbox} 
\usepackage{textcomp}
\usepackage{xcolor}
\usepackage{color}
\usepackage{tabularray}
\def\BibTeX{{\rm B\kern-.05em{\sc i\kern-.025em b}\kern-.08em
    T\kern-.1667em\lower.7ex\hbox{E}\kern-.125emX}}

\title{Data-Driven Calibration of Prediction Sets in Large Vision-Language Models Based on Inductive Conformal Prediction}

\author{ {Yuanchang Ye} \\
	School of Data Sciences\\
	Zhejiang University of Finance \& Economics\\
	HangZhou, China \\
	\And
	{Yanwen Wei} \\
	School of Data Sciences\\
	Zhejiang University of Finance \& Economics\\
	HangZhou, China \\
}

\renewcommand{\shorttitle}{IEEE 3rd International Conference on Image 
Processing and Computer Applications}

\hypersetup{
pdftitle={A template for the arxiv style},
pdfsubject={q-bio.NC, q-bio.QM},
pdfauthor={David S.~Hippocampus, Elias D.~Striatum},
pdfkeywords={First keyword, Second keyword, More},
}

\begin{document}
\maketitle

\begin{abstract}
	This study addresses the critical challenge of hallucination mitigation in Large Vision-Language Models (LVLMs) for Visual Question Answering (VQA) tasks through a \textbf{Split Conformal Prediction (SCP)} framework. While LVLMs excel in multi-modal reasoning, their outputs often exhibit hallucinated content with high confidence, posing risks in safety-critical applications. We propose a model-agnostic uncertainty quantification method that integrates dynamic threshold calibration and cross-modal consistency verification. By partitioning data into calibration and test sets, the framework computes \textit{nonconformity scores} to construct prediction sets with statistical guarantees under user-defined risk levels ($\alpha$). Key innovations include: (1) rigorous control of \textbf{marginal coverage} to ensure empirical error rates remain strictly below $\alpha$; (2) dynamic adjustment of prediction set sizes inversely with $\alpha$, filtering low-confidence outputs; (3) elimination of prior distribution assumptions and retraining requirements. Evaluations on benchmarks (ScienceQA, MMMU) with eight LVLMs demonstrate that SCP enforces theoretical guarantees across all $\alpha$ values. The framework achieves stable performance across varying calibration-to-test split ratios, underscoring its robustness for real-world deployment in healthcare, autonomous systems, and other safety-sensitive domains. This work bridges the gap between theoretical reliability and practical applicability in multi-modal AI systems, offering a scalable solution for hallucination detection and uncertainty-aware decision-making.
\end{abstract}


\section{Introduction}
With the rapid advancement of multi-modal models~\cite{bi2025prism}, large vision-language models (LVLMs)~\cite{bi2024visual} have been widely deployed in critical sectors such as healthcare and autonomous driving~\cite{kostumov2024uncertainty,zhang2024vl,liu2023visual,zhang2023spot,yang2025lighthouse}. However, research on vision-language question answering (VQA) tasks indicates that, compared to uni-modal language models, these multi-modal systems are more susceptible to pronounced hallucination phenomena~\cite{rohrbach2018object,rawte2023survey}. Despite generating responses that often appear convincing and exhibiting high confidence, the models can produce inaccurate outputs. Relying on such hallucinated results may introduce decision-making biases or even pose significant safety risks. In this context, developing efficient and automated hallucination detection mechanisms has emerged as a central challenge to ensuring the reliability of multi-modal AI systems. Moreover, studies show that processing visual and textual information together in VQA tasks increases the risk of hallucinations. These issues highlight the need for automated detection frameworks that can adapt to multi-modal uncertainty without relying on prior knowledge. Our approach, which integrates dynamic threshold calibration and cross-modal consistency validation, aims to offer real-time, robust reliability in safety-sensitive applications.

Prior research has focused on quantifying model outputs and providing users with measures to assess reliability in natural language generation (NLG)~\cite{liang2024survey,li2023evaluating}. Current uncertainty quantification methods, such as calibration-based techniques and verbalized uncertainty approaches, aim to signal prediction trustworthiness. However, these methods—often heuristic in nature—fail to deliver task-specific performance guarantees, limiting their practical applicability. For example, verbalized uncertainty frequently exhibits overconfidence, undermining its reliability. While calibration aligns probabilities with empirical correctness rates, it requires costly retraining and remains vulnerable to distribution shifts. These limitations highlight the need for more robust and generalizable frameworks to ensure trustworthy uncertainty estimation in NLG.

\textbf{Conformal Prediction }(CP) is an uncertainty quantification framework whose primary advantage lies in providing rigorous statistical guarantees on the coverage of true outcomes based solely on the data exchangeability assumption~\cite{romano2019conformalized,cresswell2024conformal,ke2025statistical}. In contrast to methods that rely on heuristic approximations or complex prior distributions, CP is model-agnostic, distribution-free, and computationally efficient, allowing it to be directly applied to pretrained systems without the need for retraining. In this work, we adopt the Split Conformal Prediction (SCP) method and extend it to the multi-choice scenario in closed-ended Vision-VQA tasks. Specifically, the candidate answer set for the target dataset is first generated using LVLMs, and then, based on the calibration set samples’ ground-truth labels, a Nonconformity Score (NS) is designed to quantify the uncertainty of the model outputs. By computing the quantile of the NS in the calibration set and incorporating a user-specified risk level (denoted as $\delta$), strict control over the marginal coverage is ultimately achieved on the test set. This method not only avoids the dependency on distributional assumptions inherent to traditional approaches, but also provides theoretical support for reliable decision-making in multi-modal scenarios.
\par Our experiments utilize MMMU and ScienceQA as benchmark datasets and evaluate eight LVLMs from four distinct model groups, including LLaVA1.5, LLaVA-NeXT, Qwen2VL, and InternVL2. Extensive empirical results demonstrate that our framework achieves rigorous control over the miscoverage rate across various user-specified risk levels (denoted as $\alpha$). For instance, on the ScienceQA benchmark, even with a high tolerance for error probability ($\alpha \geq 0.6$), the Qwen2-VL-7B-lnstruct model maintains its empirical error rate below $\alpha = 0.6$. Notably, as $\alpha$ increases, the average prediction size of generated answer sets systematically tightens—a critical property for mitigating hallucination in LVLMs. This inverse relationship between $\alpha$ and prediction set size ensures that higher risk tolerance yields more compact prediction sets, effectively filtering out low-confidence or spurious outputs. Furthermore, regardless of the calibration-to-test data split ratio, the average empirical error rate consistently adheres to the user-defined risk level. Combined with the controllable prediction set granularity, this robustness underscores the method's dual capability: ensuring statistically valid coverage while dynamically suppressing hallucinatory responses through adaptive set constraints. Such capabilities are pivotal for deploying LVLMs in safety-critical scenarios where both reliability and precision are paramount.

\section{Related Work}
\textbf{Large Vision-Language Models.} Early-stage research primarily focused on generating text responses from image and text inputs. Building upon this foundation, subsequent studies have significantly expanded the capabilities and application domains of LVLMs. Recent advances further enhance fine-grained parsing abilities, enabling precise control over localized regions (e.g., bounding boxes or key points) beyond holistic image understanding. These developments have facilitated the widespread deployment of LVLMs in critical fields such as medical diagnosis, embodied robotics interaction, and autonomous driving. However, the complexity of multimodal interactions introduces new challenges—for instance, inconsistencies in cross-modal information fusion may degrade output reliability. In high-risk scenarios like healthcare and autonomous systems, unreliable model responses could lead to severe safety hazards, underscoring the necessity of accurate hallucination detection. Unlike traditional methods reliant on external validation, this work proposes quantifying the intrinsic uncertainty of LVLMs to identify hallucinations, establishing a novel theoretical foundation for building safe and dependable human-AI collaborative systems.

\textbf{Hallucinations in Large Language Models.} In natural language processing~\cite{yang2025selfgoal}, hallucination refers to generated content that appears plausible but deviates from source material or factual accuracy, originating from psychological concepts of perceiving non-existent realities~\cite{lin2023generating,kuhn2023semantic,farquhar2024detecting,wang2025word,yang2024logu}. This phenomenon manifests primarily as two types: intrinsic hallucination (direct contradictions with source context) and extrinsic hallucination (content unverifiable through original context or external knowledge bases). Research on Large Vision-Language Models (LVLMs) reveals that their strong focus on user-centric interaction and instruction alignment leads to factual distortions, categorized as factual hallucination (divergence from verifiable truths) and faithfulness hallucination (violations of user instructions, contextual coherence, or logical consistency) .Detection methodologies follow two approaches. (1) External-model-based evaluation: This approach employs advanced LVLMs as scoring discriminators to assess response quality, though constrained by reliance on synthetic annotations. (2) Discrete rule-based checking: discrete rule-based systems focus on Object Hallucination (OH) assessment through benchmarks like CHAIR, MME, and POPE. Mitigation strategies employ Contrastive Decoding (CD) and post-processing techniques: CD addresses perceptual biases through visual region comparisons, self-contrast analysis, and preference model comparisons, yet suffers from sensitivity and oversimplification; post-processing optimizes responses via iterative prompting but faces computational overhead and limited task adaptability. This framework provides multidimensional insights for systematically evaluating LVLM output reliability.

\textbf{Split Conformal Prediction (SCP).} SCP demonstrates unique strengths as a theoretically grounded uncertainty quantification framework for Large Vision-Language Models (LVLMs). Its core mechanism leverages exchangeable data calibration to generate prediction sets with statistical guarantees of covering ground-truth answers, applicable to black-box models handling open-ended natural language generation tasks~\cite{campos2024conformal,angelopoulos2023conformal,wang2024conu,ye2024benchmarking,angelopoulos2024conformal,wang2024sample,wang2025sconu}. Unlike conventional uncertainty frameworks, SCP requires minimal assumptions while delivering verifiable coverage guarantees. The method remains model-agnostic and distribution-free, operating solely under exchangeable data conditions. Recent extensions adapt SCP to multimodal scenarios through dynamic prediction set construction using confidence thresholds (e.g., candidate answer filtering in QA tasks) or likelihood-based stopping rules for generation sequences. Addressing limitations in open-ended generation, advanced implementations deploy black-box uncertainty quantification strategies that rigorously link uncertainty metrics to correctness criteria, enabling robust coverage guarantees across diverse model architectures and data complexities. Despite challenges like non-exchangeable data adaptation and real-time computational demands, SCP's model independence, distribution-free nature, and bias-control capabilities establish it as a theoretically rigorous and practically viable solution for assessing LVLM output reliability.

\section{Method}
Our method primarily addresses two challenges. (1) How to identify the distribution of responses within the model’s output that satisfy user requirements. (2) How to rigorously demonstrate that the identified output distribution meets the model’s statistical guarantees. We first developed an uncertainty quantification method based on non-conformity scores to establish a reliability measure for model-generated responses. Furthermore, we employ Split Conformal Prediction to systematically transform heuristic approximations of uncertainty quantification results into statistically rigorous ones. This approach ensures both the robustness of prediction sets and stronger statistical guarantees, thereby providing theoretically grounded assurances for the model’s output distribution.
\subsection{Preliminaries}
Consider a prediction task where \(\mathcal{X}\) and \(\mathcal{Y}\) denote the input and output sets, respectively. Following previous research on the CP framework, we first establish a Calibration Set comprising \(n\) samples, denoted by \(\{(X_i, Y_i)\}_{i=1}^{n}\). Next, for the \(K\)-class classification task, we collect \(M\) samples as test data, denoted by \(\{(X_i, Y_i)\}_{i=N+1}^{M+N}\). Thereafter, we define an LVLM classifier model \(\hat{f} : \mathcal{X} \rightarrow \mathcal{Y}\). We denote the correct class of the \(i\)th sample (ground truth) as \(y_i^*\).

For each data point, without any system prompt processing, the sample is directly fed into the LVLM classifier \(\hat{f}\), and \(P\) random samplings are performed. The results obtained for the \(i\)-th data sample are denoted by \(\hat{y}^{(i)}_k\); for instance, when \(i = 1\) and \(K = 5\), the random sampling results are 
\[
\{\underbrace{\hat{y}^{(1)}_1, \hat{y}^{(1)}_2, \hat{y}^{(1)}_2, \hat{y}^{(1)}_3, \ldots, \hat{y}^{(1)}_5, \hat{y}^{(1)}_5}_{P}\}.
\]

Within the CP framework, a key component is the \textbf{nonconformity score} \(\mathcal{S} : \mathcal{X} \times \mathcal{Y} \rightarrow \mathbb{R}\), which provides a heuristic measure of how well the classifier’s prediction conforms to the given input. For classification tasks, the classifier’s output distribution is employed as 
\begin{equation}
\mathcal{S}(x, y) = 1 - \hat{f}(y \mid x) \quad \text{for } y = 1, \ldots, |\mathcal{Y}|,
\end{equation}
we denote \(S_i = \mathcal{S}(X_i, Y_i)\) as the nonconformity score for the \(i\)-th calibration example.
\subsection{Method}
For a new, unseen test sample \( x_{test} \), the steps to generate \(\hat{C}_{\alpha}(X_{test})\) are as follows:

\begin{enumerate}
\item Compute the nonconformity scores for the calibration data, \((S_1, \dots, S_N)\), where \( S_i = \mathcal{S}(X_i, Y_i) \);
\item Define \(\tau = Q_{1-\alpha}(\{S_i\}_{i=1}^{n})\) as the conformal \(\alpha\) quantile of the empirical distribution of the scores, where \(\alpha \in (0,1)\) is the significance level we set. A smaller value of \(\alpha\) corresponds to a lower allowable error rate. In computing \(\tau\), the set \(\{S_i\}_{i=1}^{N}\) is arranged in descending order, and the quantile corresponding to the \(\lceil (1-\alpha)(n+1)/n \rceil\)-th order statistic is selected;
\item Finally, we use the previously defined prediction set:
    \begin{equation}
    \hat{C}_{\alpha}(x_{\text{test}}) = \left\{ y \in \mathcal{Y} : \mathcal{S}(x_{\text{test}}, y) \leq \tau \right\}
    \end{equation}
\end{enumerate}

Steps 1 and 2 are commonly referred to as \textbf{calibration}, while step 3 is known as \textbf{prediction}. Intuitively, the prediction set includes all predictions corresponding to samples that conform at least as well as a sufficiently large portion of the calibration set.

\subsection{Theoretical Guarantees}
The coverage guarantee of conformal prediction (CP) stems from its two fundamental theoretical properties: \textbf{distribution-free validity} and \textbf{marginal coverage}. As demonstrated by Vovk et al. (2005), the prediction sets generated by the conformal predictor defined in the previous subsection satisfy the following coverage guarantee:
\begin{equation}
\mathbb{P}\left[Y_{\text{test}} \in \hat{C}_{\alpha}(X_{\text{test}})\right] \geq 1 - \alpha
\end{equation}
provided the data satisfies \textit{exchangeability}. Exchangeability requires that the joint probability distribution of the data remains invariant under permutations. Formally, a sequence \((Z_1, \dots, Z_n)\) is exchangeable if and only if for any permutation \(\pi\) of \(\{1, \dots, n\}\):
\[
(Z_1, \dots, Z_n) \triangleq (Z_{\pi(1)}, \dots, Z_{\pi(n)})
\]
where \(\triangleq\) denotes equality in distribution. Exchangeability is a weaker condition than independence and identical distribution (i.i.d.). While i.i.d. variables are necessarily exchangeable, exchangeable variables need not be independent—they need only be identically distributed. This yields the Romano upper bound:
\begin{equation}
\mathbb{P}\{Y_{n+1} \in \hat{C}(X_{n+1})\} \leq 1 - \alpha + \frac{1}{n+1}.
\end{equation}

Notably, as the calibration set size \(n\) increases, the coverage probability converges strictly to \(1 - \alpha\). Crucially, the described conformal prediction procedure exhibits \textbf{model-agnosticism} and \textbf{distribution-free validity}—it makes no assumptions about the data distribution beyond exchangeability.In our subsequent tasks, we should use this formula to evaluate whether the results meet our guarantees:
\begin{equation}
\mathbb{P}\left[Y_{\mathrm{test}}\notin\mathcal{C}_{\alpha}(X_{\mathrm{test}})\right]\leq\alpha
\end{equation}
When the \textbf{Empirical Error Rate} is lower than the $\alpha$ level, we can confidently conclude that the prediction set satisfies the specified coverage guarantees.

\section{Evaluations}
\subsection{ Experimental Set-up}
\textbf{Benchmarks.} Our experiments utilize multiple-choice benchmarks. For the multiple-choice datasets, we employ two benchmarks: MMMU and ScienceQA. Specifically, MMMU contains 11.5K multimodal questions from university-level 

\begin{table}[ht]
\centering
\resizebox{\textwidth}{!}{
\begin{tblr}{
  row{1} = {c},
  row{2} = {c},
  row{3} = {c},
  row{4} = {c},
  row{5} = {c},
  row{6} = {c},
  row{7} = {c},
  row{8} = {c},
  row{9} = {c},
  row{10} = {c},
  row{12} = {c},
  row{13} = {c},
  row{14} = {c},
  row{15} = {c},
  row{16} = {c},
  row{17} = {c},
  row{18} = {c},
  row{19} = {c},
  cell{1}{1} = {r=2}{},
  cell{1}{2} = {r=2}{},
  cell{1}{3} = {c=9}{},
  cell{3}{1} = {r=8}{},
  cell{11}{1} = {c=2}{c},
  cell{12}{1} = {r=8}{},
  cell{20}{1} = {c=2}{c},
  vline{2-3} = {1-10,12-19}{},
  vline{3} = {4-11,13-120}{},
  vline{2} = {11,20}{},
  hline{1,3,11-12,20-21} = {-}{},
  hline{2} = {3-11}{},
}
Benchmarks & LVLMs                    & Split\_Ratio &         &         &         &         &         &         &         &         \\
           &                          & 0.1          & 0.2     & 0.3     & 0.4     & 0.5     & 0.6     & 0.7     & 0.8     & 0.9     \\
ScienceQA  & Qwen2-VL-7B-Instruct     & \textbf{0.1895}       & 0.1949  & 0.1943  & 0.1955  & 0.1951  & 0.1947  & 0.1923  & 0.1921  & 0.1908  \\
           & Qwen2-VL-2B-Instruct     & \textbf{0.1827}       & 0.1879  & 0.1908  & 0.1851  & 0.1959  & 0.1888  & 0.1927  & 0.1950  & 0.1827  \\
           & InternVL2-8B             & \textbf{0.1847}       & 0.1901  & 0.1984  & 0.1942  & 0.1936  & 0.1937  & 0.1955  & 0.1959  & 0.1975  \\
           & InternVL2-1B             & 0.1874       & \textbf{0.1869}  & 0.1932  & 0.1895  & 0.1905  & 0.1899  & 0.1901  & 0.1873  & 0.1937  \\
           & llava-v1.6-vicuna-13b-hf & \textbf{0.1810}       & 0.1887  & 0.1923  & 0.1865  & 0.1912  & 0.1909  & 0.1930  & 0.1833  & 0.1902  \\
           & llava-v1.6-mistral-7b-hf & \textbf{0.1825}       & 0.1847  & 0.1871  & 0.1841  & 0.1922  & 0.1923  & 0.1869  & 0.1884  & 0.1880  \\
           & llava-1.5-13b-hf         & 0.1889       & 0.1905  & \textbf{0.1844}  & 0.1952  & 0.1959  & 0.1918  & 0.1917  & 0.1909  & 0.1960  \\
           & llava-1.5-7b-hf          & 0.1902       & \textbf{0.1867}  & 0.1892  & 0.1948  & 0.1925  & 0.1921  & 0.1879  & 0.1900  & 0.1876  \\
\textbf{Average}   &                          & \textbf{0.1859}~      & 0.1888~ & 0.1912~ & 0.1906~ & 0.1934~ & 0.1918~ & 0.1913~ & 0.1904~ & 0.1908~ \\
MMMU       & Qwen2-VL-7B-Instruct     & \textbf{0.1579}       & 0.1750  & 0.1858  & 0.1828  & 0.1829  & 0.1795  & 0.1879  & 0.1880  & 0.1970  \\
           & Qwen2-VL-2B-Instruct     & 0.1640       & 0.1651  & \textbf{0.1632}  & 0.1758  & 0.1789  & 0.1848  & 0.1821  & 0.1877  & 0.1924  \\
           & InternVL2-8B             & \textbf{0.1740}       & 0.1799  & 0.1800  & 0.1824  & 0.1843  & 0.1745  & 0.1854  & 0.1823  & 0.1783  \\
           & InternVL2-1B             & \textbf{0.1423}       & 0.1538  & 0.1637  & 0.1706  & 0.1673  & 0.1605  & 0.1699  & 0.1619  & 0.1510  \\
           & llava-v1.6-vicuna-13b-hf & \textbf{0.1463}       & 0.1577  & 0.1614  & 0.1673  & 0.1626  & 0.1579  & 0.1576  & 0.1580  & 0.1531  \\
           & llava-v1.6-mistral-7b-hf & \textbf{0.1617}       & 0.1713  & 0.1780  & 0.1821  & 0.1770  & 0.1857  & 0.1788  & 0.1775  & 0.1853  \\
           & llava-1.5-13b-hf         & \textbf{0.1578}       & 0.1747  & 0.1765  & 0.1790  & 0.1835  & 0.1844  & 0.1888  & 0.1900  & 0.1903  \\
           & llava-1.5-7b-hf          & 0.1731       & 0.1719  & 0.1758  & 0.1777  & 0.1735  & 0.1878  & 0.1774  & 0.1782  & \textbf{0.1669}  \\
\textbf{Average}    &                          & \textbf{0.1596}~      & 0.1687~ & 0.1731~ & 0.1772~ & 0.1763~ & 0.1769~ & 0.1785~ & 0.1779~ & 0.1768~ 
\end{tblr}
}
\caption{Under the fixed $\alpha = 0.2$, we compare the error rate $\alpha$ across different split ratios on two benchmarks and eight LVLMs. First, whether in the smaller calibration set or in the broader calibration set, the $\alpha$ values remain below our preset $\alpha = 0.2$. Moreover, by observing the average $\alpha$ value at each split ratio, we find that despite using only a small number of samples, the test set results are quite good. This demonstrates that our SCP method can guarantee the test set results.
}
\end{table}

\noindent exams, quizzes, and textbooks, covering six core disciplines: Art \& Design, Business, Science, Health \& Medicine, Humanities \& Social Science, and Technology \& Engineering. These questions span 30 subjects and 183 subfields, involving 30 types of highly heterogeneous images. MMMU also provides a complete test set with 150 development samples and 900 validation samples. For ScienceQA, the questions originate from open resources managed by IXL Learning, an online educational platform curated by K-12 domain experts. This dataset includes questions aligned with the California Common Core Content Standards, containing 21,208 samples split into training (12,726), validation (4,241), and test (4,241) sets.

\textbf{Base LVLMs.} In this experiment, we evaluate 8 LVLM models from 4 distinct model groups. Specifically, we conduct inference on the aforementioned benchmarks using LLaVA-1.5, LLaVA-NeXT, Qwen2-VL, and InternVL2. LLaVA-1.5 aligns the CLIP visual encoder with large language models (e.g., Vicuna) via a two-layer MLP connector, adopting a two-stage training strategy (pretraining and instruction tuning), and demonstrates strong performance in visual QA and OCR tasks. LLaVA-NeXT extends LLaVA-1.5 by introducing dynamic high-resolution processing (AnyRes), which combines global and local features through grid partitioning to enhance visual reasoning, and expands into video understanding. Qwen2-VL employs dynamic resolution adaptation to preserve fine-grained details via flexible high-resolution image splitting. InternVL2 enhances general vision-language capabilities by scaling the visual encoder (e.g., InternViT-6B), applying dynamic high-resolution processing with pixel rearrangement to reduce visual tokens, and utilizing a three-stage progressive alignment strategy.

\textbf{Implementation Details.} We implement the marginal coverage and statistical guarantees for VQA prediction sets through benchmarks, base LVLMs, and the SCP method. The detailed setup is as follows: (1) \textbf{Answer Generation Initialization.} We set the temperature of all LVLMs to 1.0 for sampling to increase answer diversity, generating 36 responses per question. (2) \textbf{Unprompted Inference.} No prompts are used; models perform inference on raw questions. Results are fed into Qwen2.5-3B-Instruct for bidirectional discrimination. Samples are discarded if no correct answer exists in the generated set. After obtaining the sampled answers, a smaller LLM checks semantic entailment between answers, converting them into fixed-length option-only sets. Semantic clustering is applied to compute cluster frequency distributions as $\hat{f}(y \mid x)$. (3) \textbf{Non-Conformity Score Generation.} Firstly, we split the results into a calibration set and a test set by \textbf{split-ratio}. We generate non-conformity scores using the method in Section III.B, averaging results over 100 rounds. Two approaches are designed: one with fixed split-ratios and another with fixed $\alpha$, measuring empirical error rates to evaluate marginal coverage and statistical guarantees.

\begin{figure}[ht]
\centering 

\begin{adjustbox}{width=\textwidth} 
\begin{tabular}{cccc} 
\subfloat[]{
    \includegraphics[width=5\textwidth]{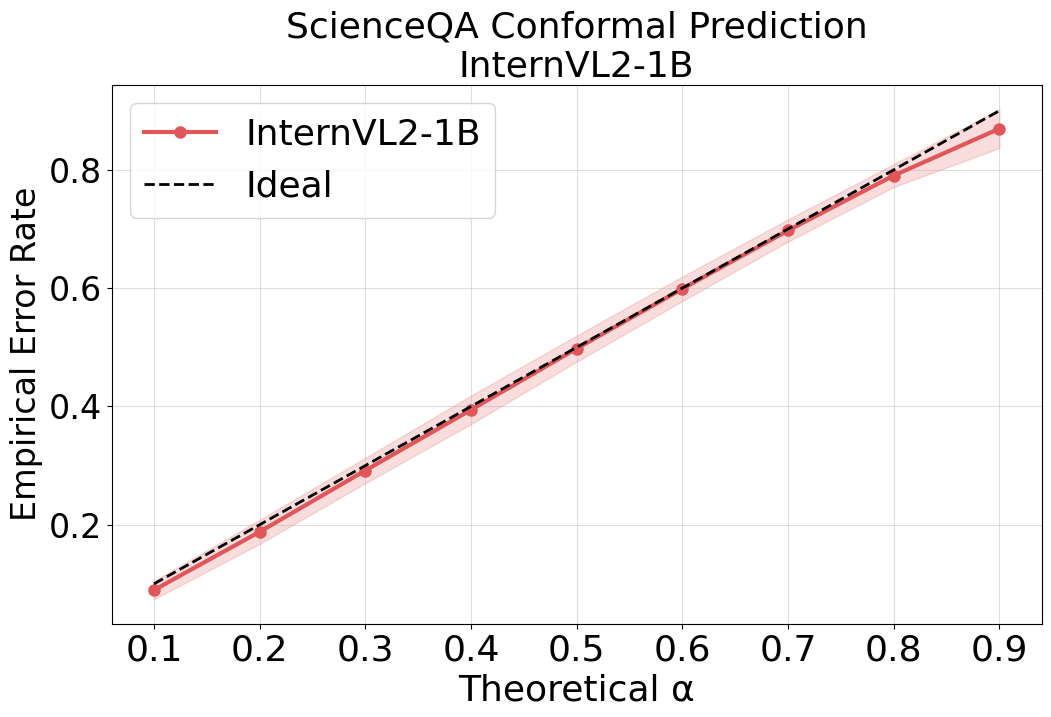}
} &
\subfloat[]{
    \includegraphics[width=5\textwidth]{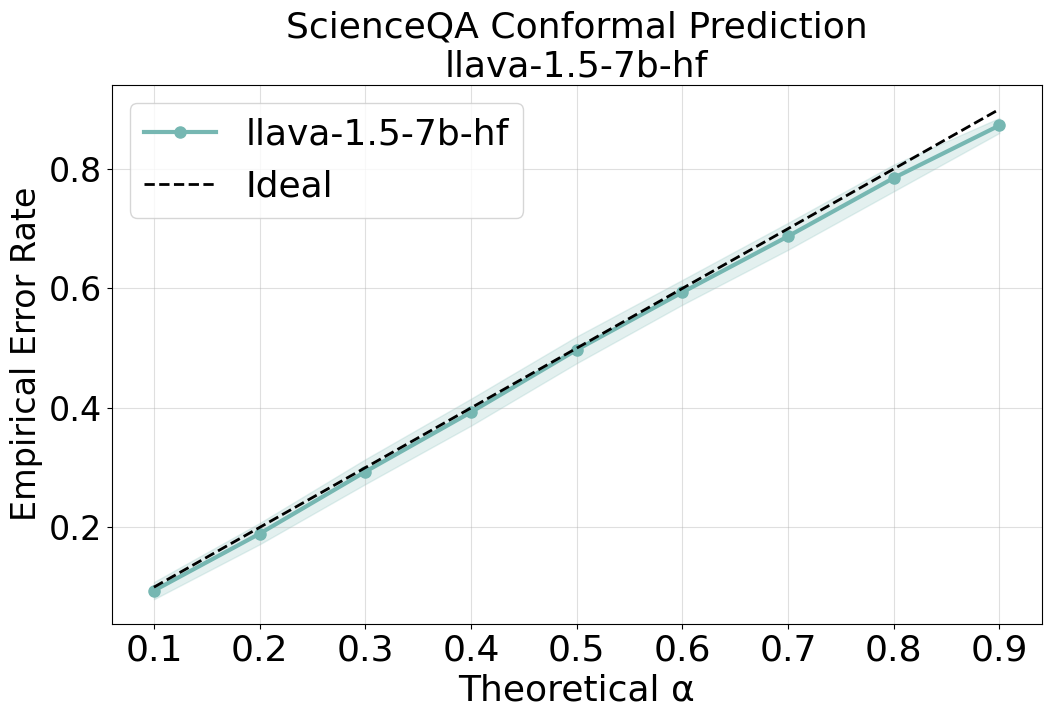}
} &
\subfloat[]{
    \includegraphics[width=5\textwidth]{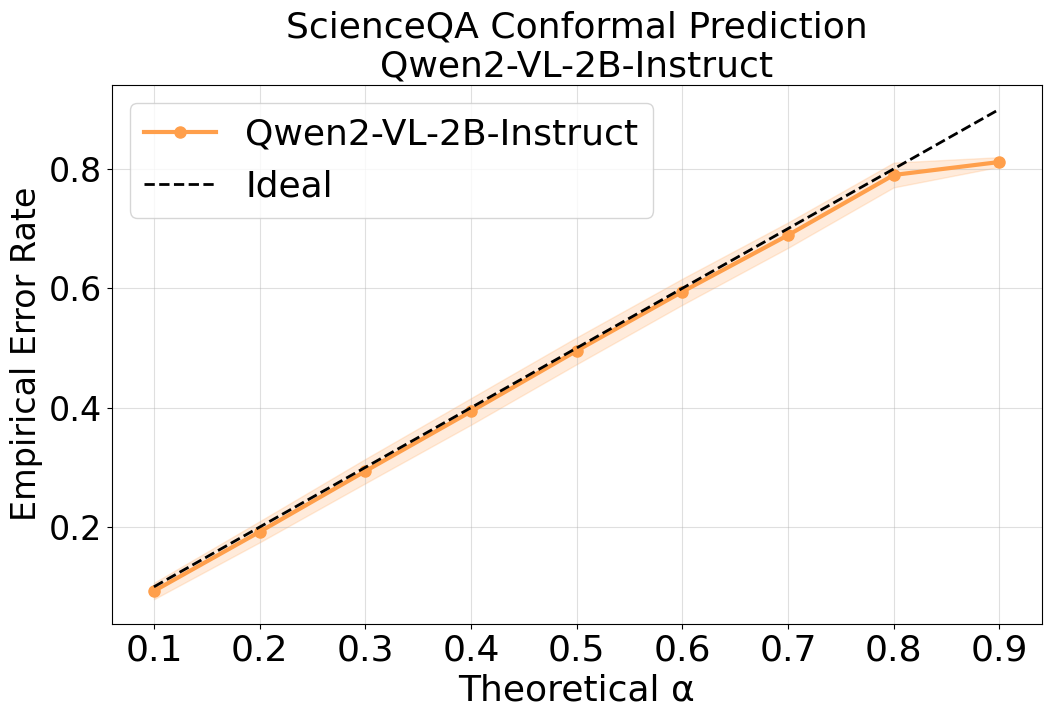}
} &
\subfloat[]{
    \includegraphics[width=5\textwidth]{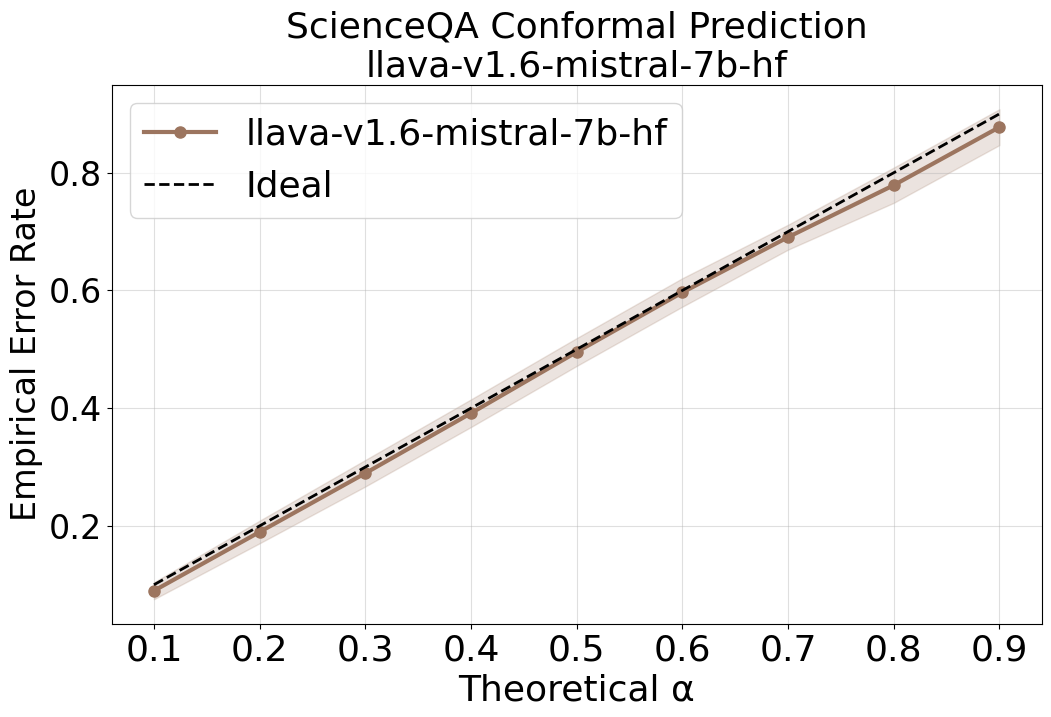}
} \\
\end{tabular}
\end{adjustbox}

\vspace{0.06cm} 

\begin{adjustbox}{width=\textwidth}
\begin{tabular}{cccc}
\subfloat[]{
    \includegraphics[width=5\textwidth]{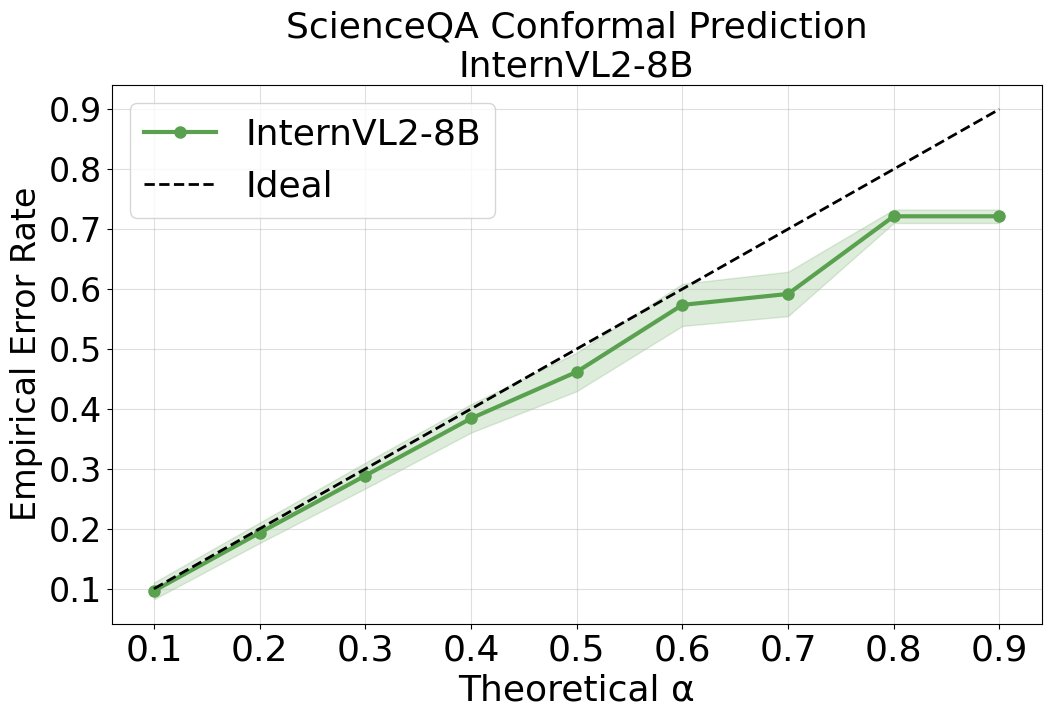}
} &
\subfloat[]{
    \includegraphics[width=5\textwidth]{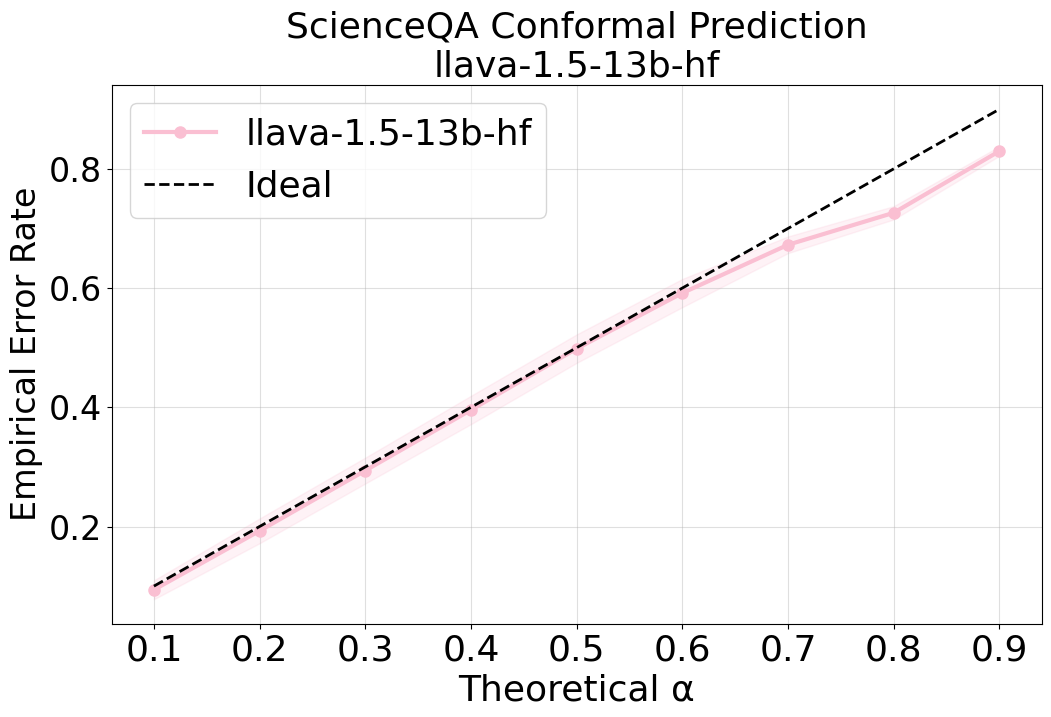}
} &
\subfloat[]{
    \includegraphics[width=5\textwidth]{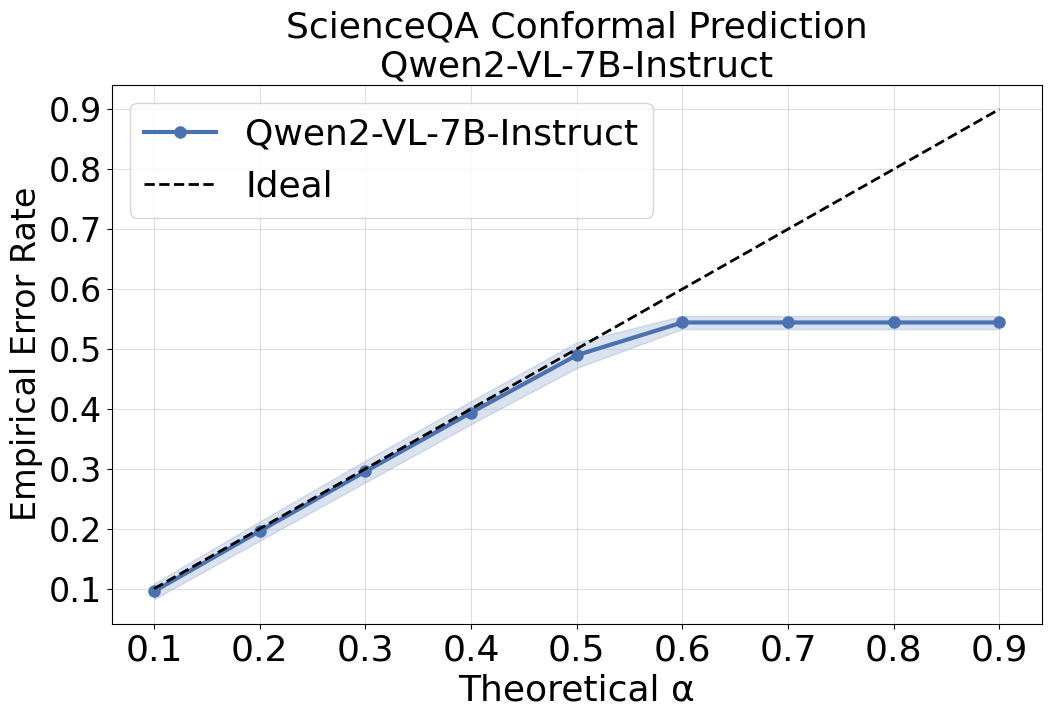}
} &
\subfloat[]{
    \includegraphics[width=5\textwidth]{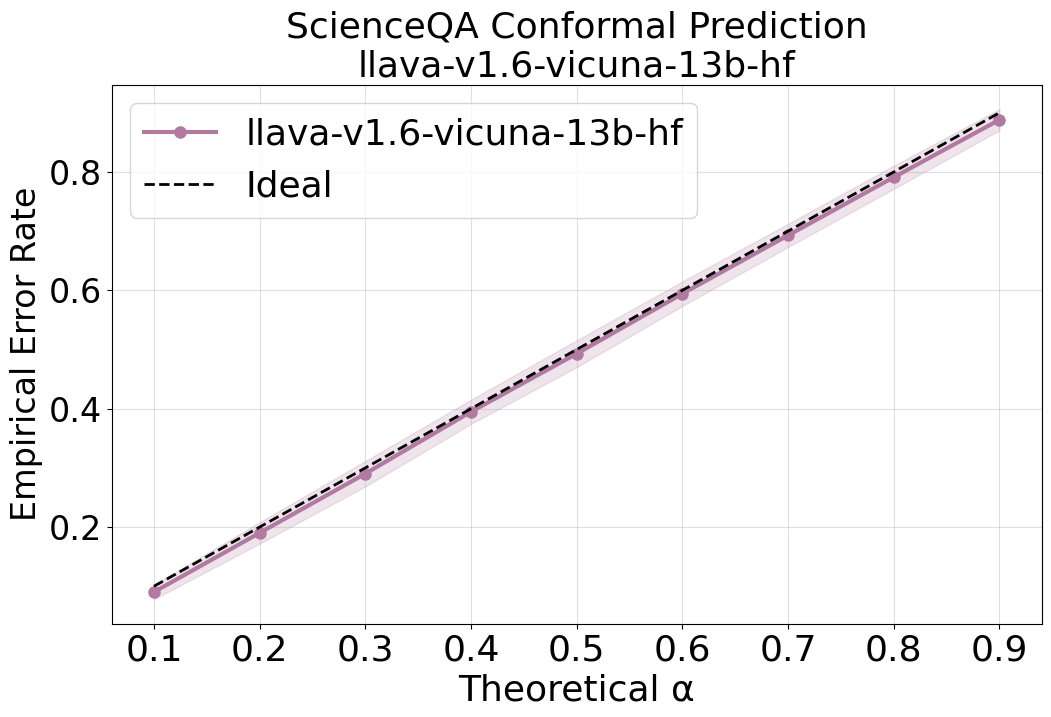}
} \\
\end{tabular}
\end{adjustbox}
\begin{adjustbox}{width=\textwidth}
\begin{tabular}{cccc}
\subfloat[]{
    \includegraphics[width=5\textwidth]{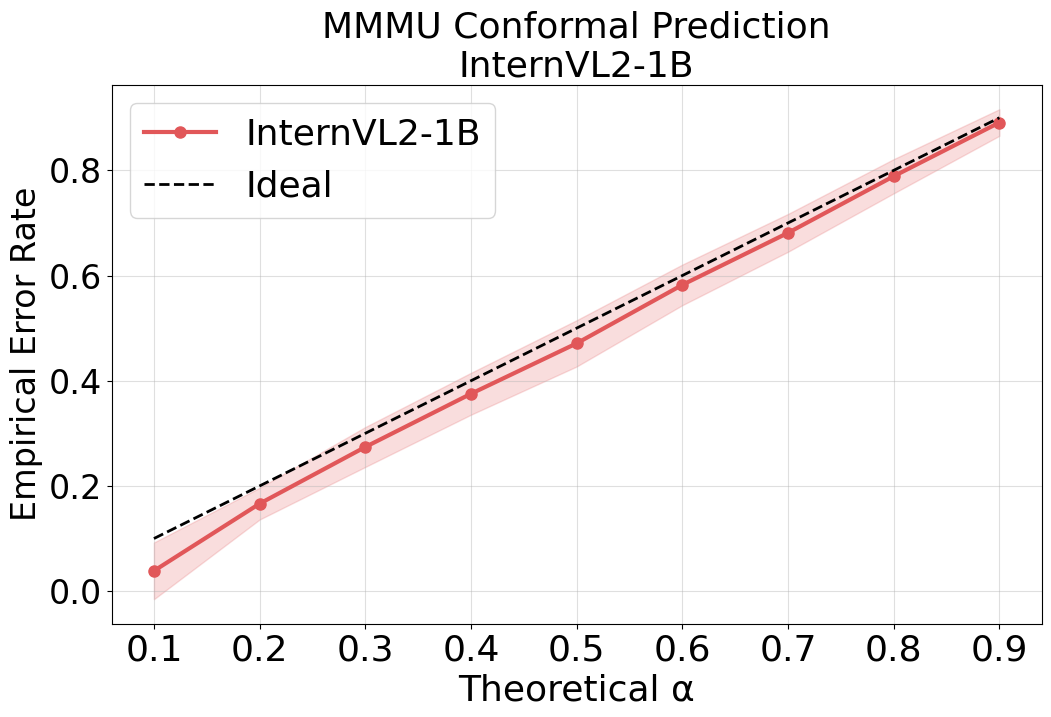}
} &
\subfloat[]{
    \includegraphics[width=5\textwidth]{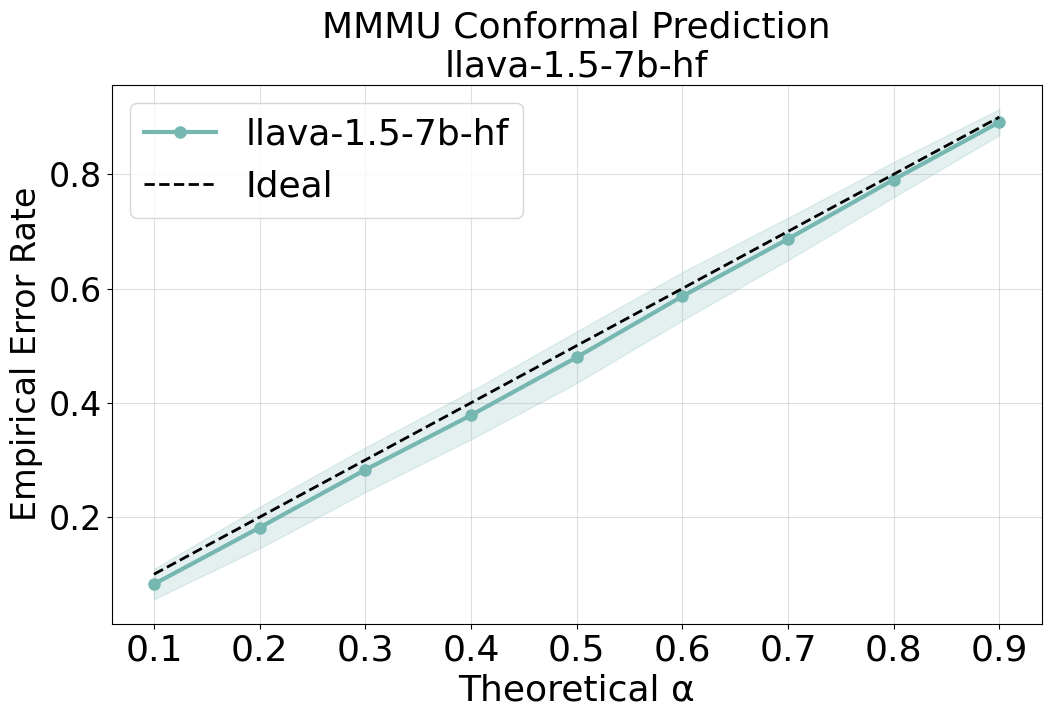}
} &
\subfloat[]{
    \includegraphics[width=5\textwidth]{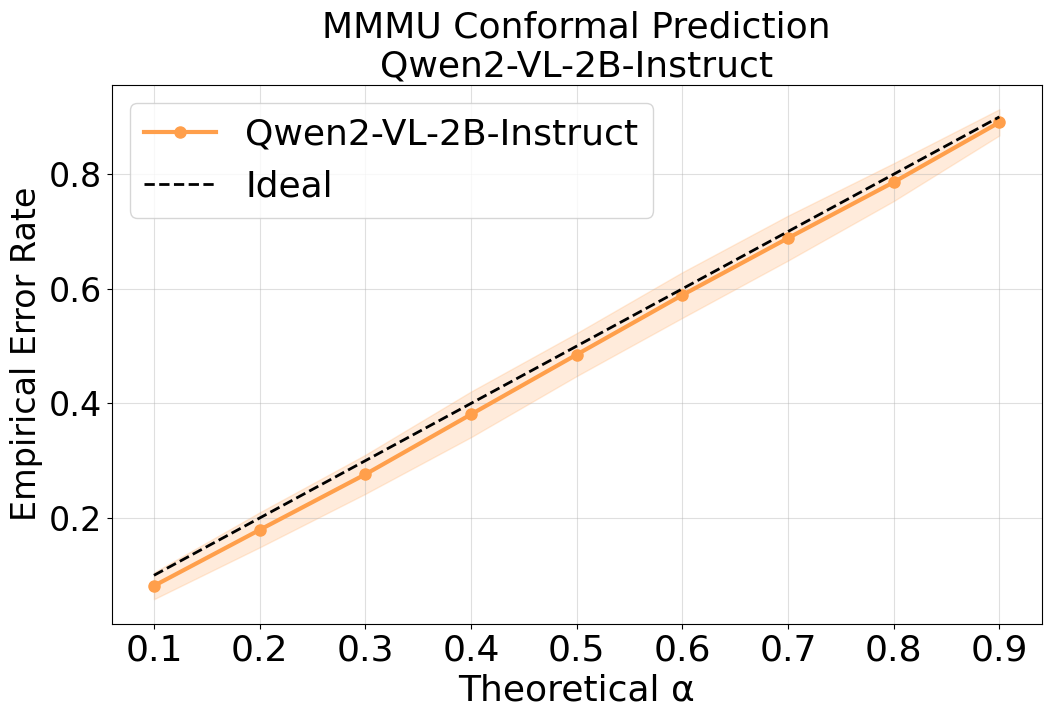}
} &
\subfloat[]{
    \includegraphics[width=5\textwidth]{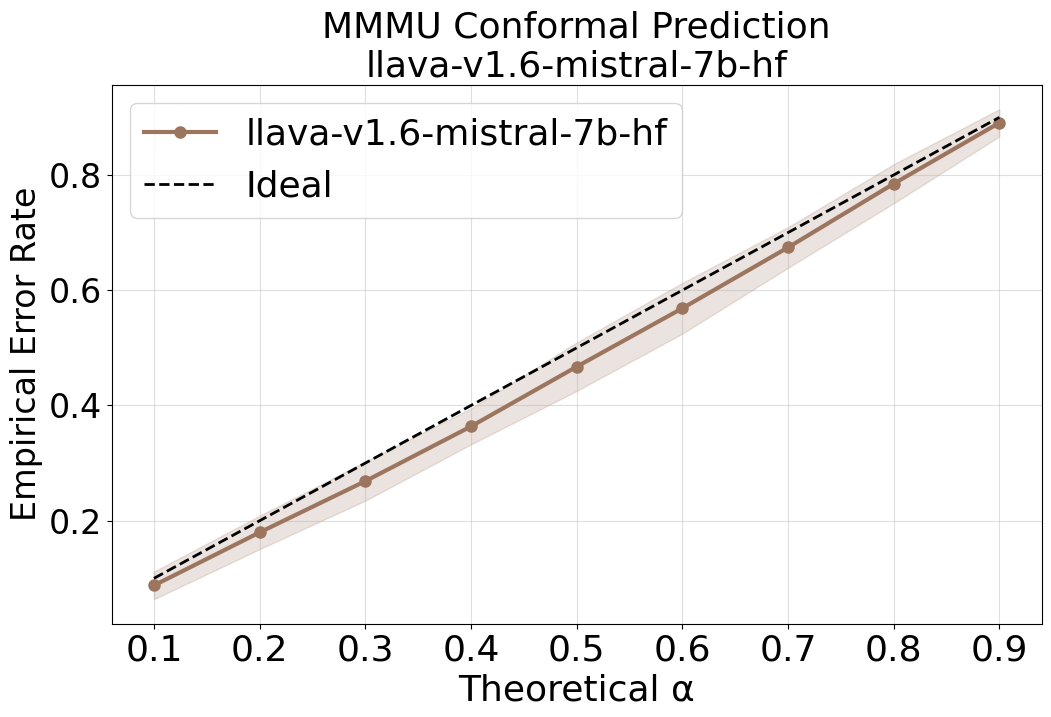}
} \\
\end{tabular}
\end{adjustbox}

\begin{adjustbox}{width=\textwidth}
\begin{tabular}{cccc}
\subfloat[]{
    \includegraphics[width=5\textwidth]{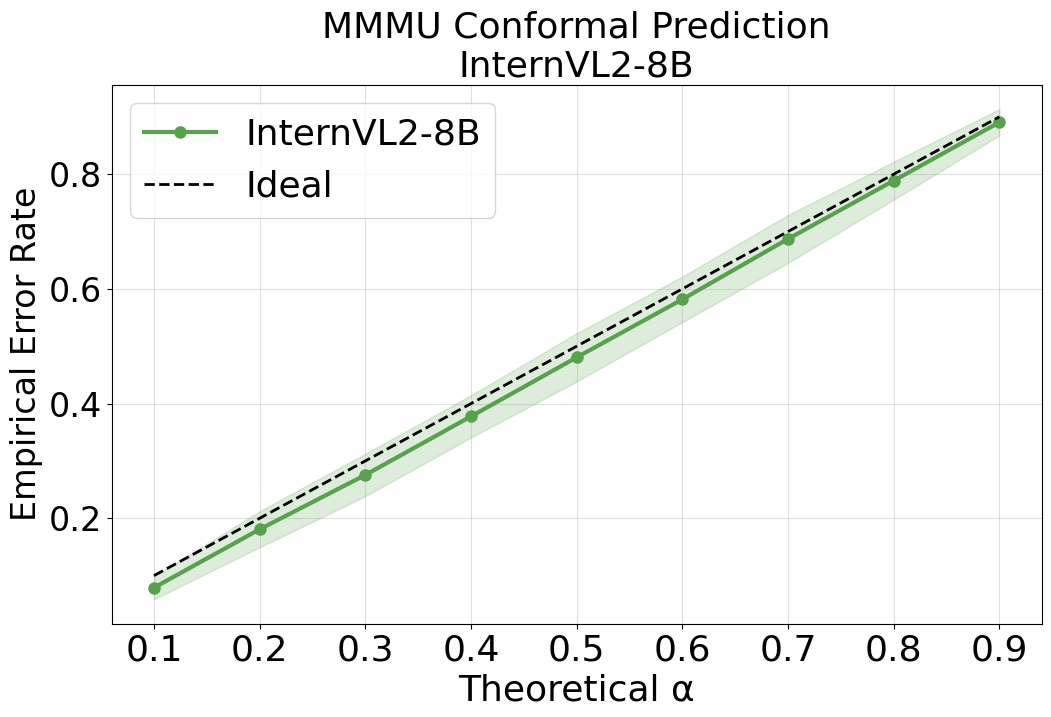}
} &
\subfloat[]{
    \includegraphics[width=5\textwidth]{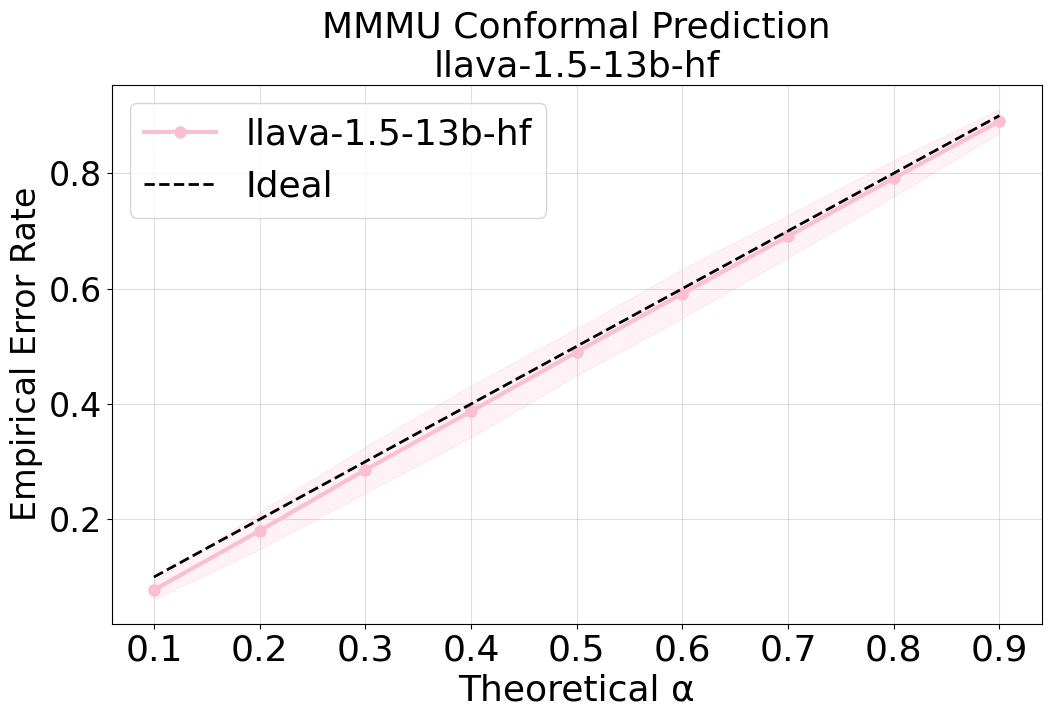}
} &
\subfloat[]{
    \includegraphics[width=5\textwidth]{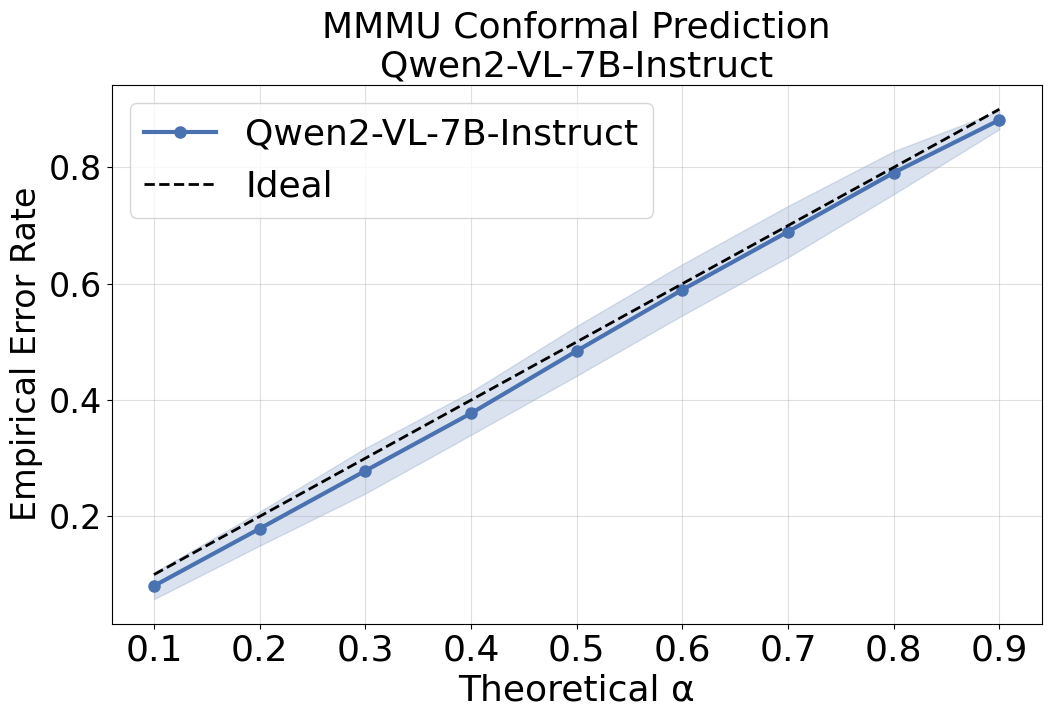}
} &
\subfloat[]{
    \includegraphics[width=5\textwidth]{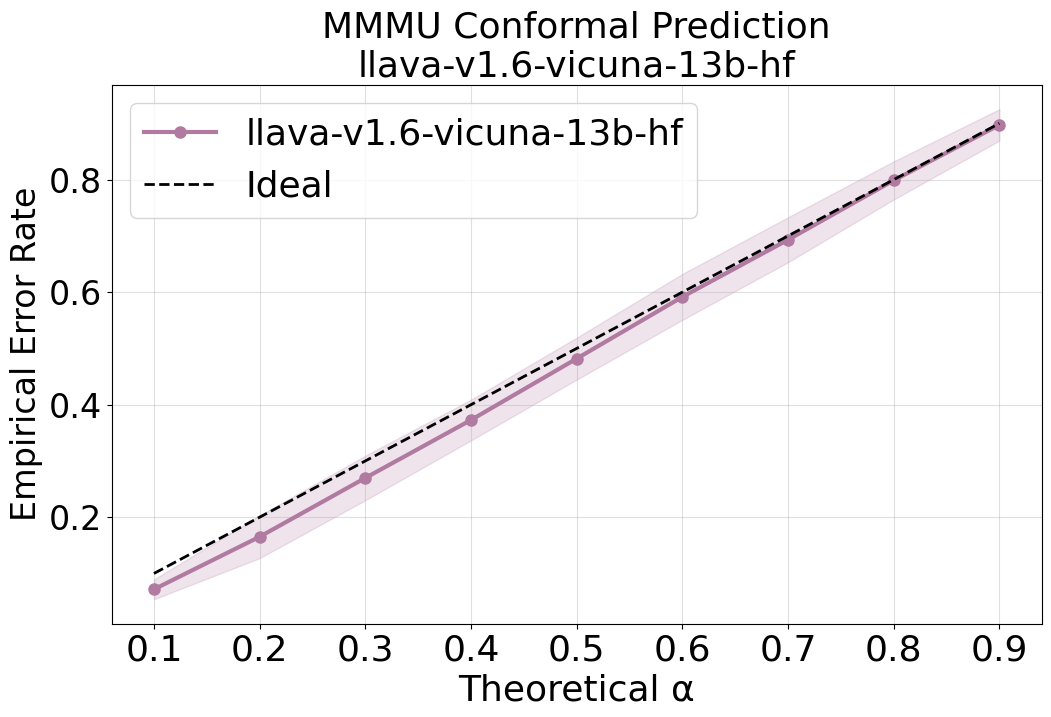}
} \\
\end{tabular}
\end{adjustbox}
\caption{Empirical Error Rate in ScienceQA and MMMU Benchmark. We calculate the mean empirical error rate and plot it as a solid line in the figure. The x-axis represents the $\alpha$ values, while the y-axis corresponds to the empirical error rate. The plot also includes a dashed line following $y = x$, indicating the ideal scenario where the empirical error rate should consistently lie below this line. Simultaneously, we compute the standard deviation of empirical error rates across 100 trials and represent it as a transparent band around the solid line.}
\label{fig:all}
\end{figure}

\subsection{Empirical Error Rate}
\textbf{Fixed Split-Ratio (0.5) Analysis.} With a fixed split-ratio of 0.5, we partitioned both datasets into 1:1 calibration and test sets. By comparing empirical error rates under varying $\alpha$, we validate that the SCP method rigorously satisfies the coverage guarantees in Equation (5) across user-specified error rates.

First, using the ScienceQA dataset (Fig.1 upper panels), the solid lines represent average empirical error rates, with transparent bands indicating variances from 100 sampling trials. As $\alpha$ increases (i.e., higher permissible error thresholds), the test error rates systematically rise while remaining strictly below $\alpha$, thereby satisfying Equation (5). Notable variations exist across LVLMs: For InternVL2-8B, empirical error rates exhibit oscillatory growth with increasing $\alpha$, whereas Qwen2-VL-7B-Instruct shows saturated error rates beyond $\alpha=0.6$. We hypothesize this saturation stems from near-optimal model outputs approaching benchmark ground truths, leaving minimal room for error escalation. This phenomenon will be further analyzed through prediction set sizes in subsequent sections.

Second, the MMMU results (Fig.1 lower panels) demonstrate superior alignment compared to ScienceQA, with lower average empirical error rates and enhanced stability despite smaller dataset size. However, MMMU exhibits larger standard deviations, likely due to its higher discriminative challenges. Crucially, both datasets maintain empirical error rates strictly below $\alpha$, reconfirming the SCP method's statistical guarantees.

\textbf{Fixed $\alpha$ with Variable Split-Ratios.} With fixed $\alpha$, we evaluate empirical error rates across split-ratios for both benchmarks and 8 LVLMs (Table 1). The results demonstrate achieved marginal coverage, consistent with the comparative analysis between MMMU and ScienceQA in prior sections.

\begin{figure}[h]
\centering 

\begin{adjustbox}{width=\textwidth} 
\begin{tabular}{cccc} 
\subfloat[]{
    \includegraphics[width=5\textwidth]{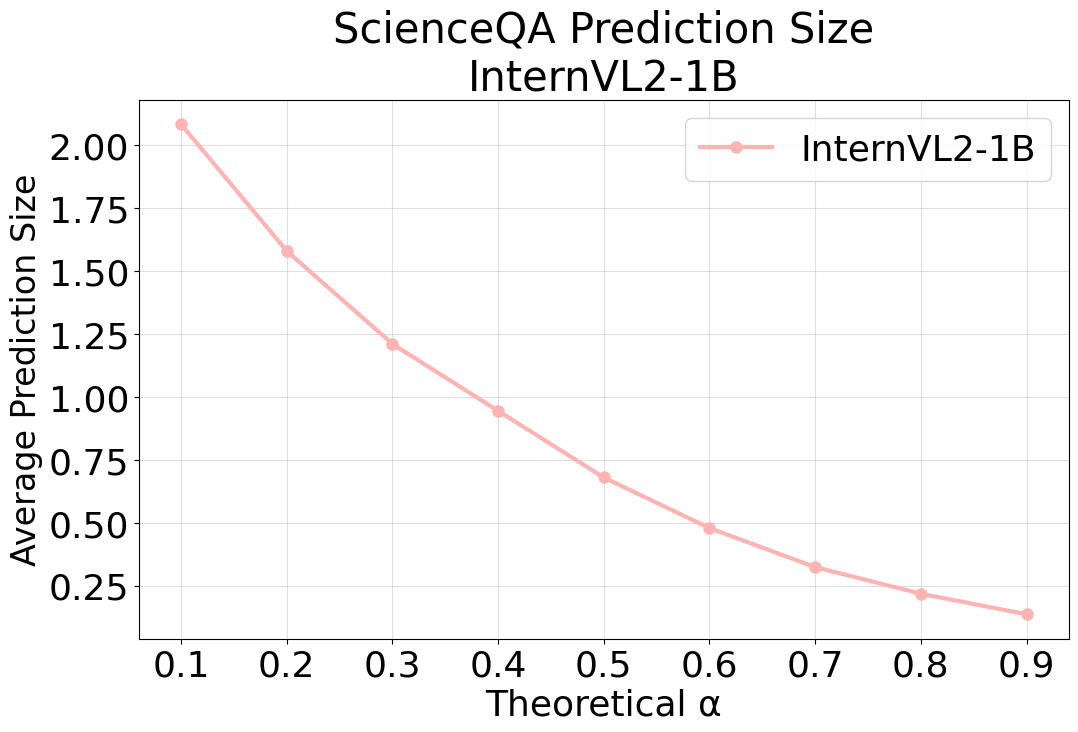}
} &
\subfloat[]{
    \includegraphics[width=5\textwidth]{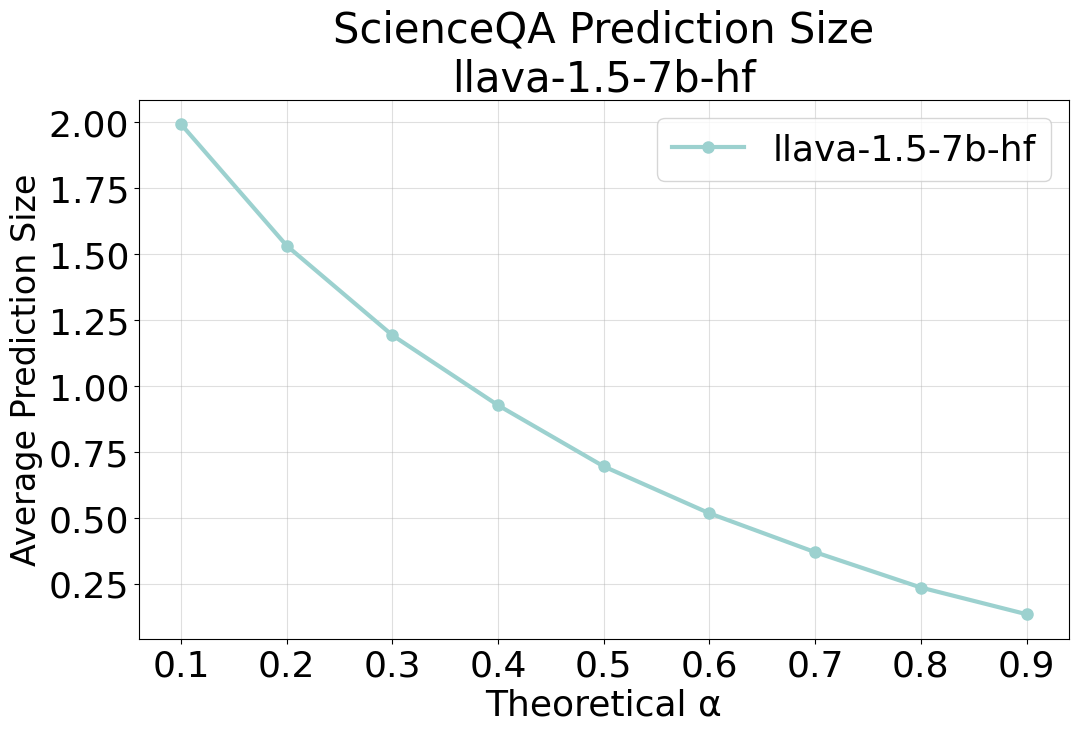}
} &
\subfloat[]{
    \includegraphics[width=5\textwidth]{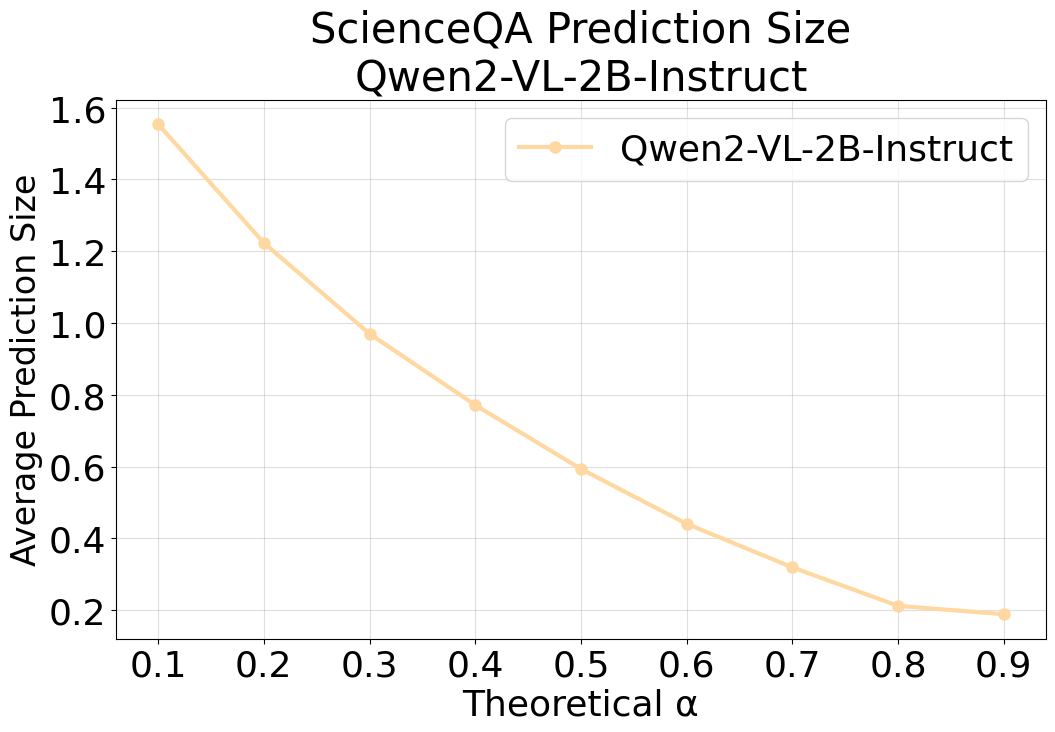}
} &
\subfloat[]{
    \includegraphics[width=5\textwidth]{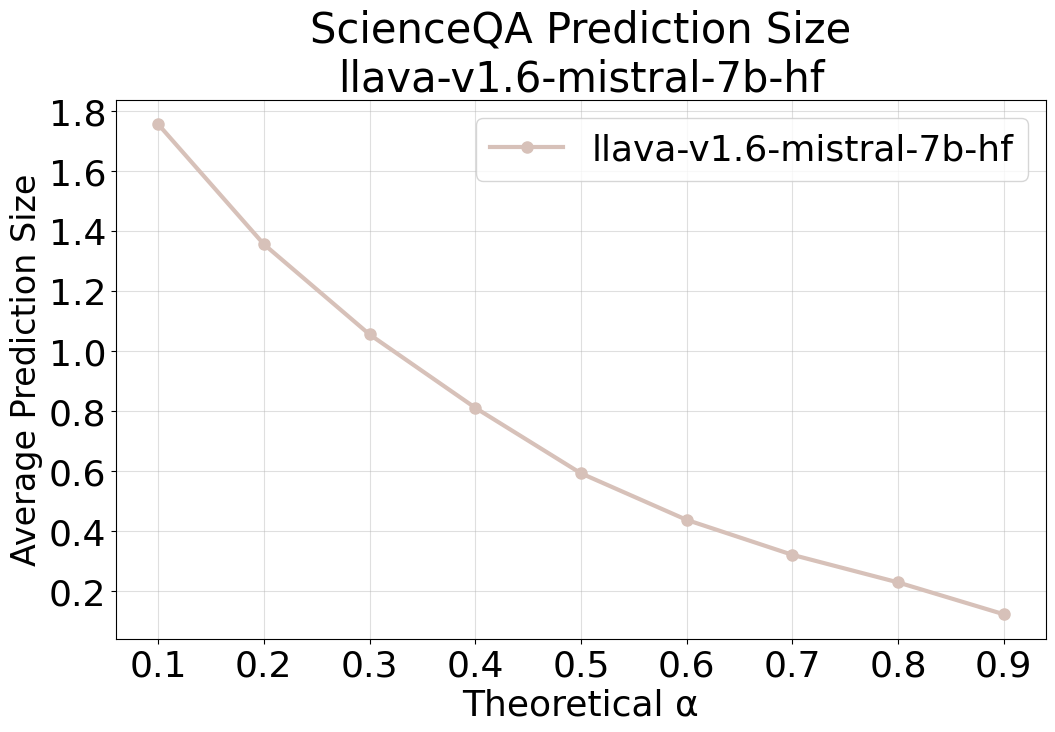}
} \\
\end{tabular}
\end{adjustbox}

\vspace{0.06cm} 

\begin{adjustbox}{width=\textwidth}
\begin{tabular}{cccc}
\subfloat[]{
    \includegraphics[width=5\textwidth]{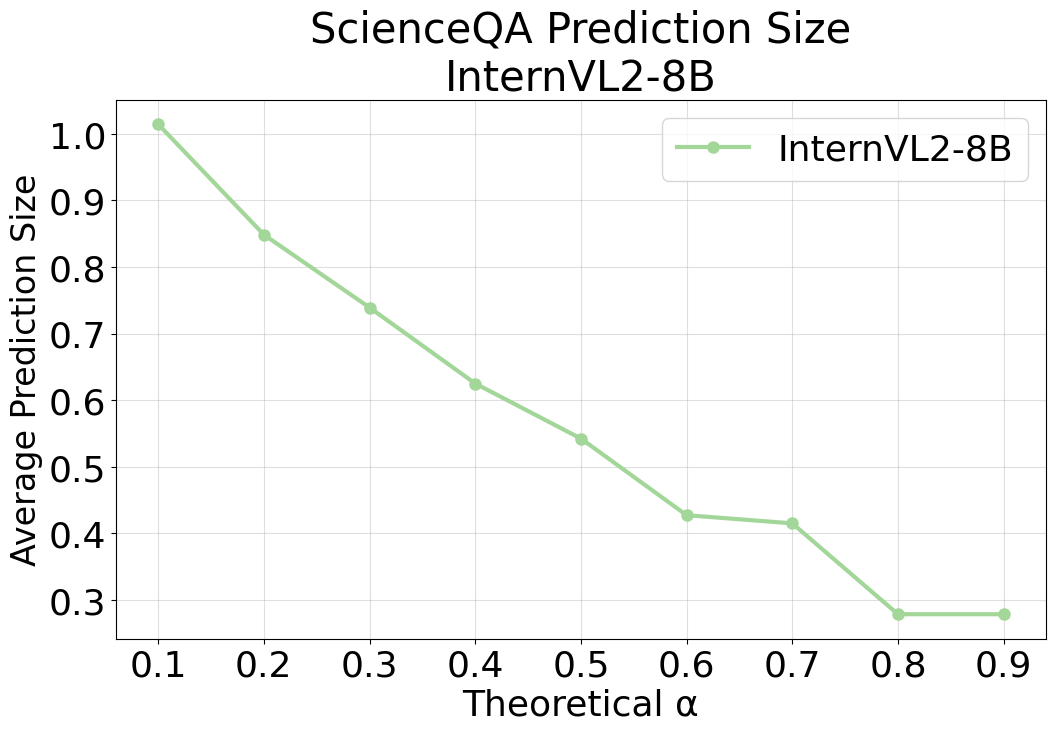}
} &
\subfloat[]{
    \includegraphics[width=5\textwidth]{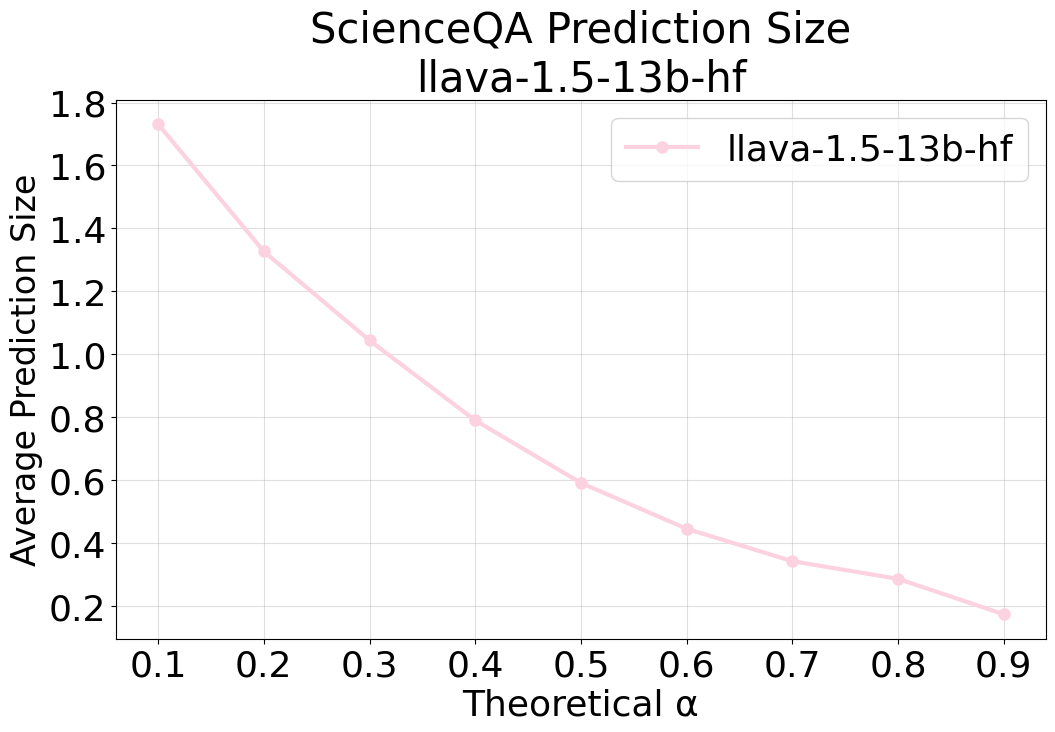}
} &
\subfloat[]{
    \includegraphics[width=5\textwidth]{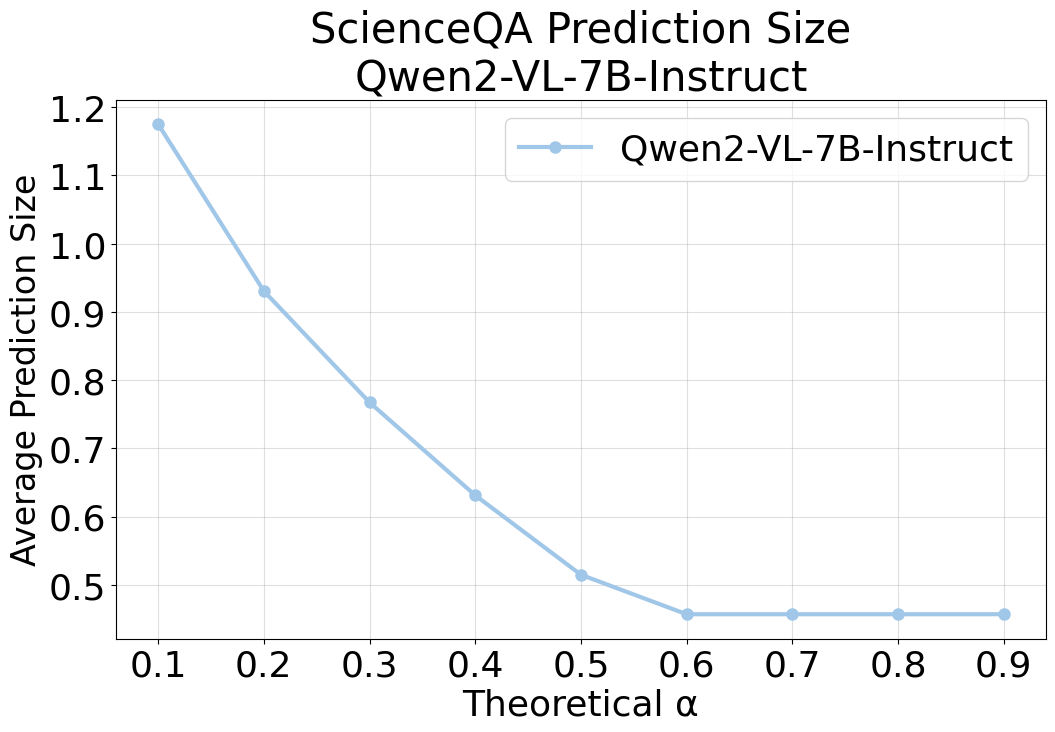}
} &
\subfloat[]{
    \includegraphics[width=5\textwidth]{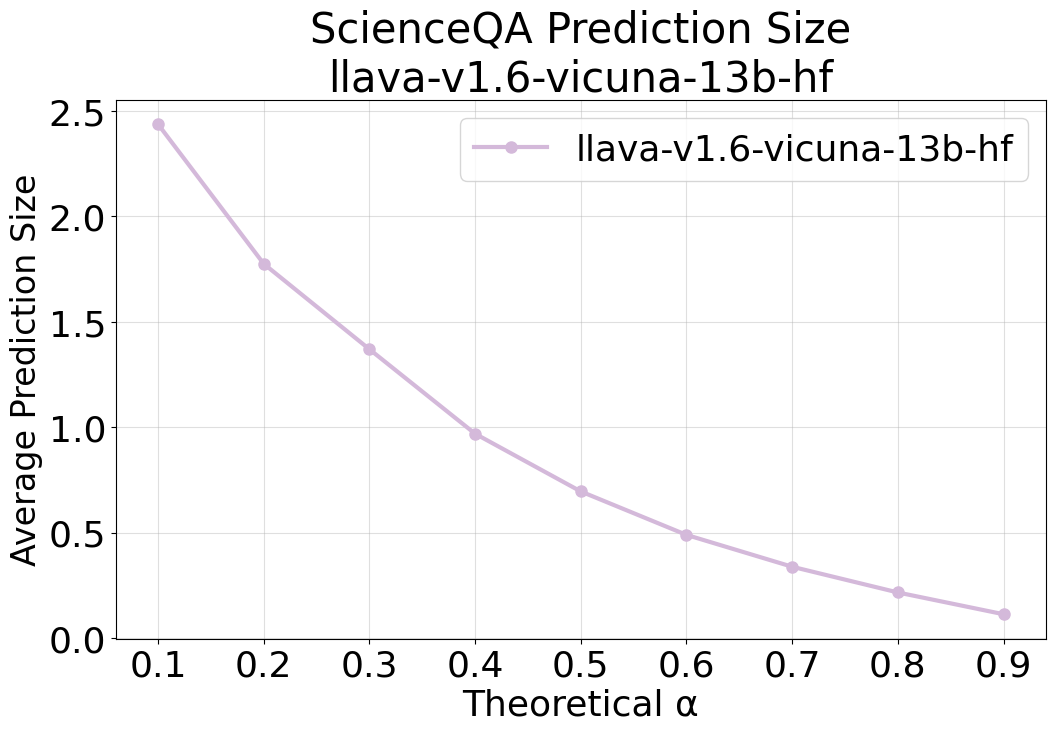}
} \\
\end{tabular}
\end{adjustbox}
\begin{adjustbox}{width=\textwidth}
\begin{tabular}{cccc}
\subfloat[]{
    \includegraphics[width=5\textwidth]{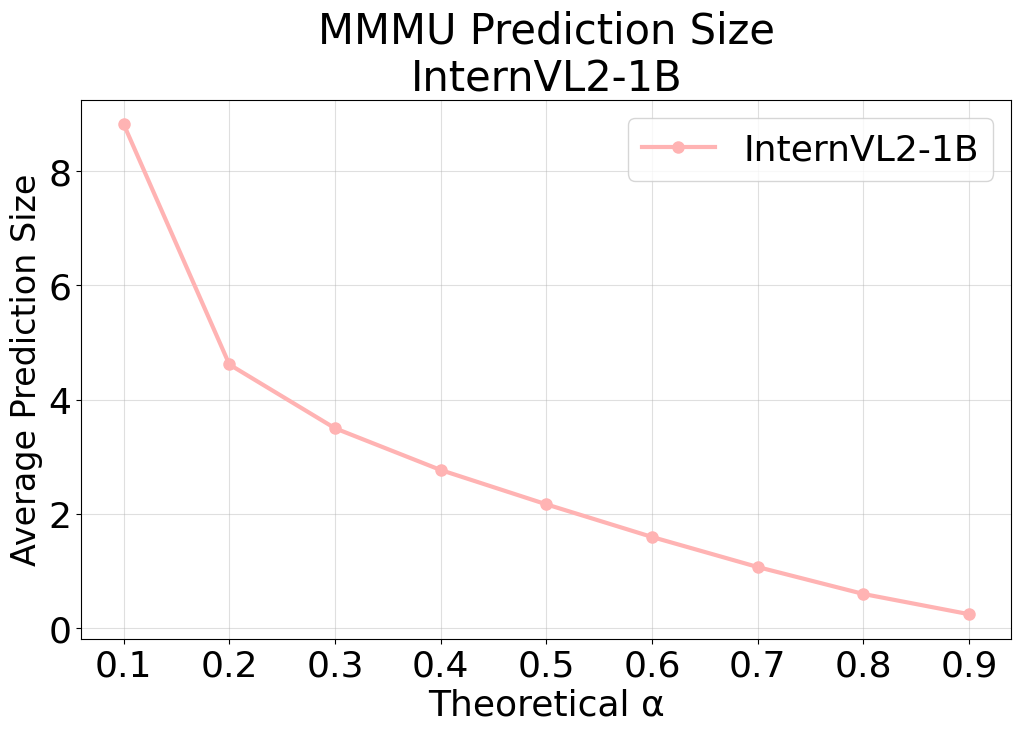}
} &
\subfloat[]{
    \includegraphics[width=5\textwidth]{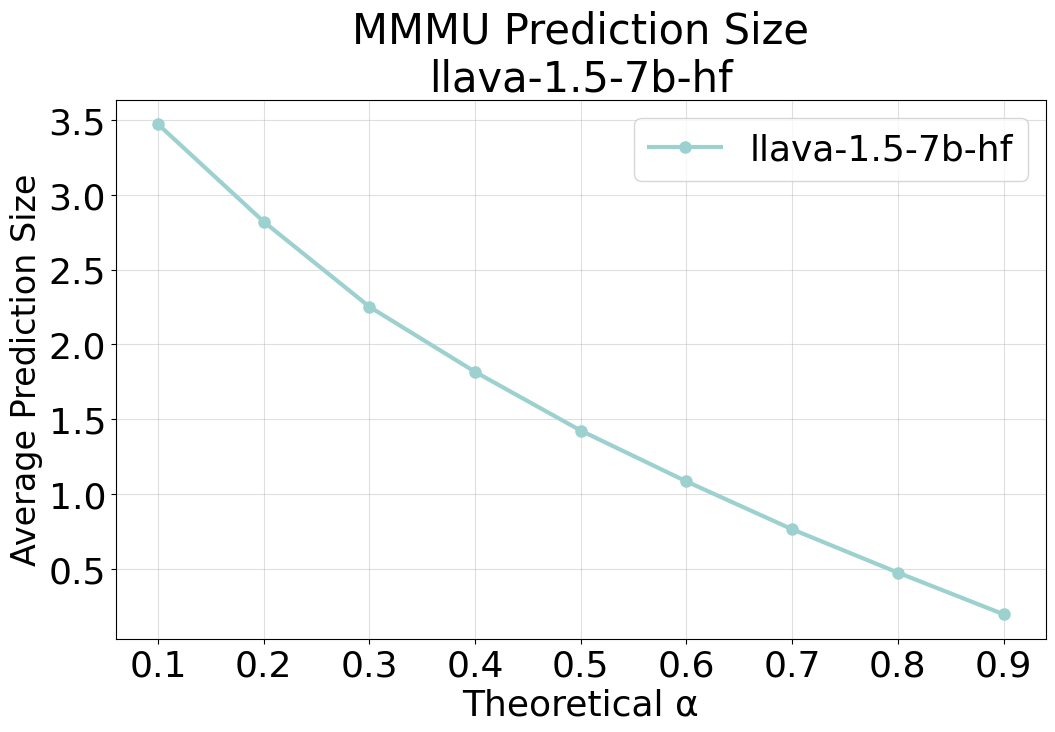}
} &
\subfloat[]{
    \includegraphics[width=5\textwidth]{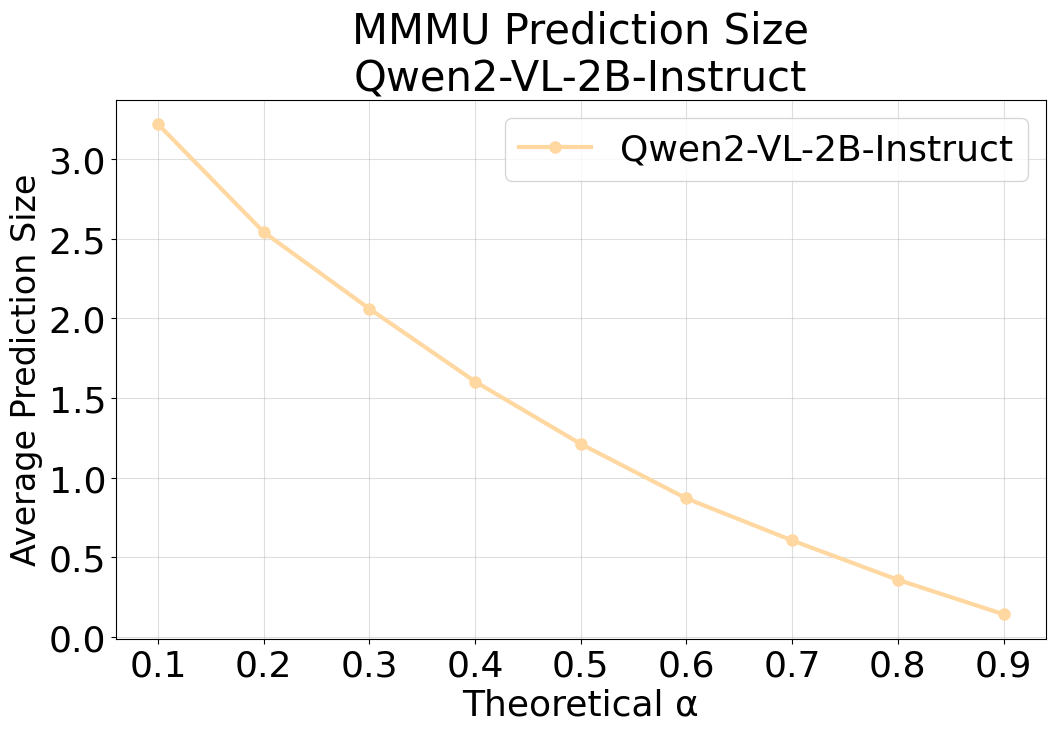}
} &
\subfloat[]{
    \includegraphics[width=5\textwidth]{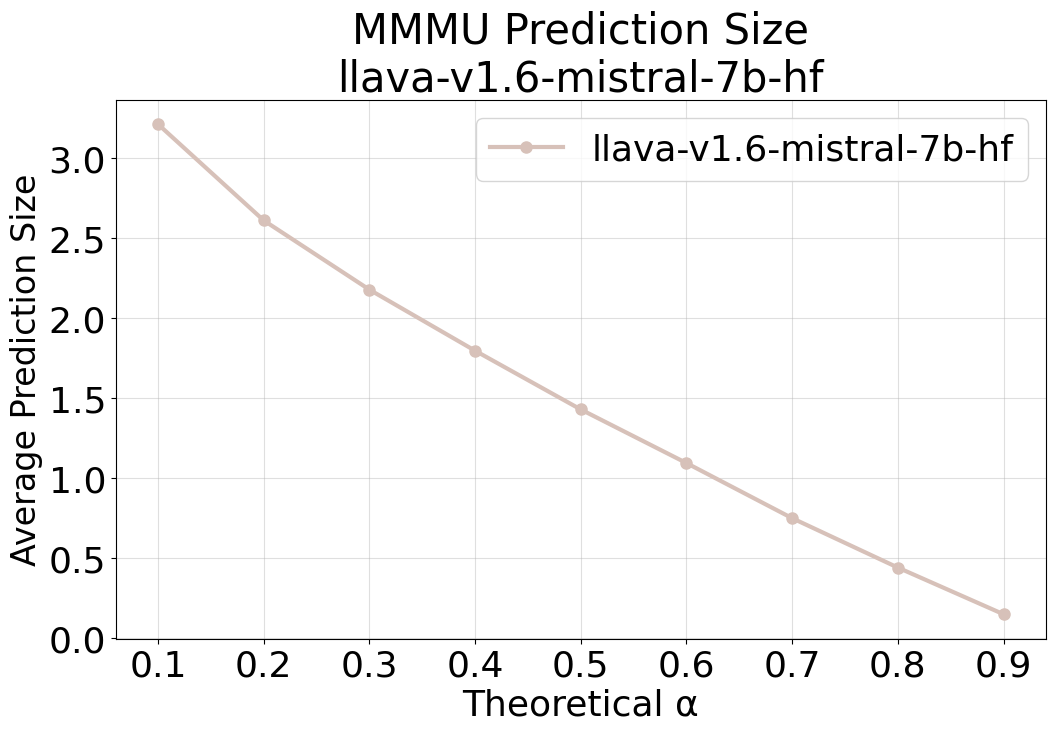}
} \\
\end{tabular}
\end{adjustbox}

\begin{adjustbox}{width=\textwidth}
\begin{tabular}{cccc}
\subfloat[]{
    \includegraphics[width=5\textwidth]{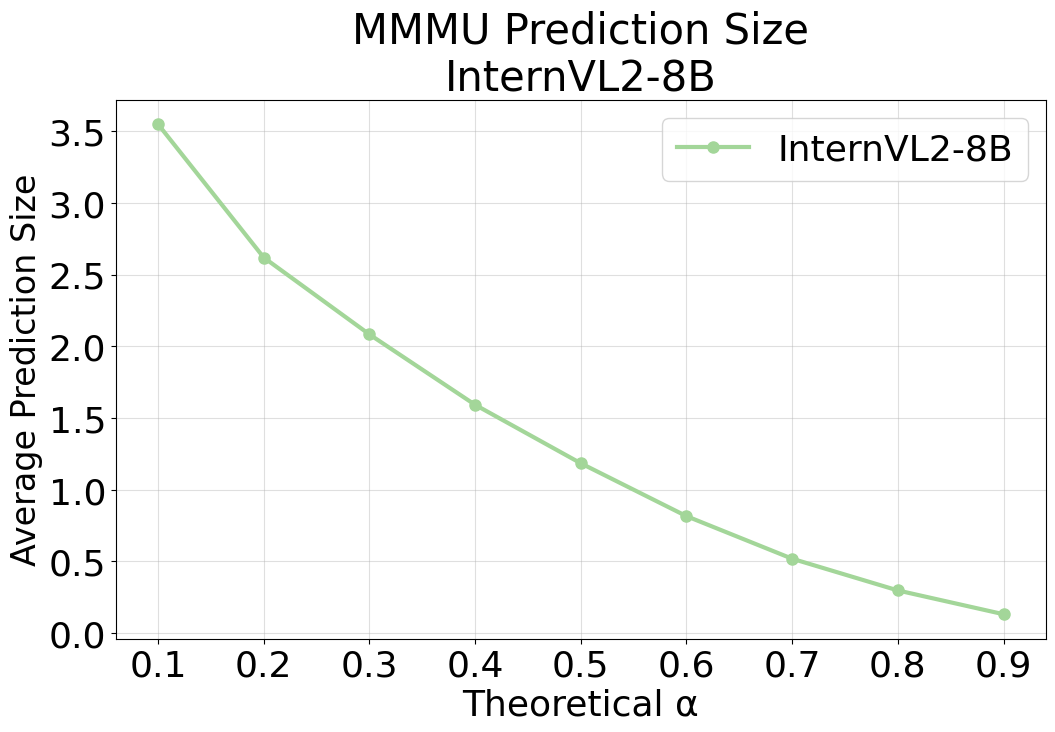}
} &
\subfloat[]{
    \includegraphics[width=5\textwidth]{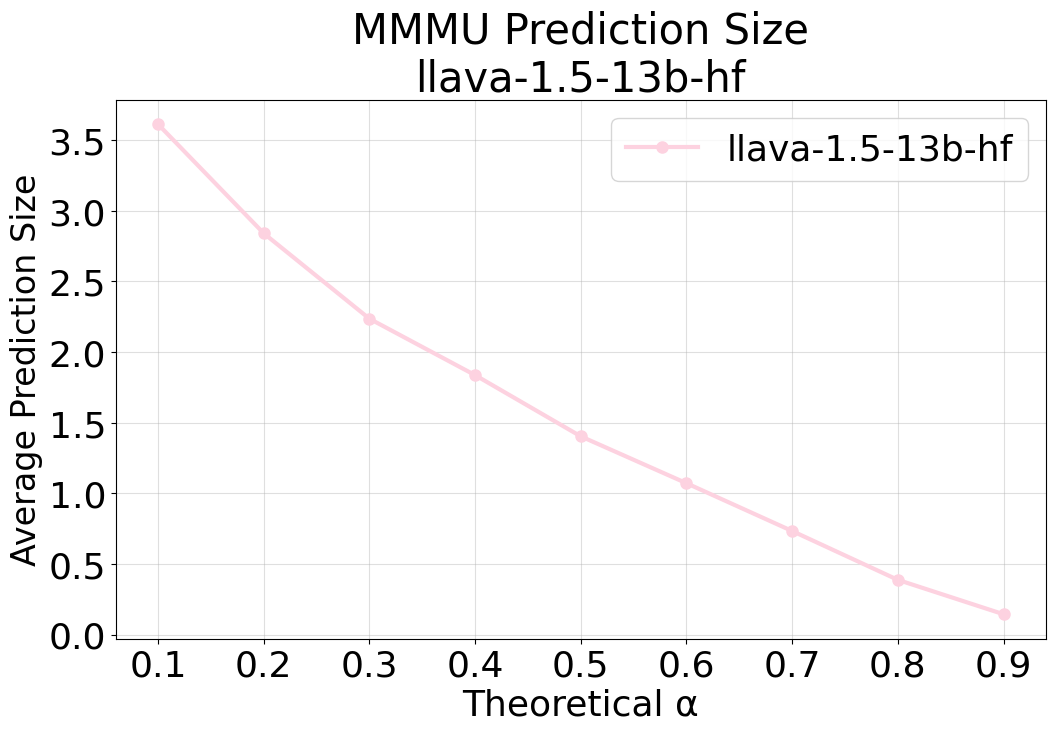}
} &
\subfloat[]{
    \includegraphics[width=5\textwidth]{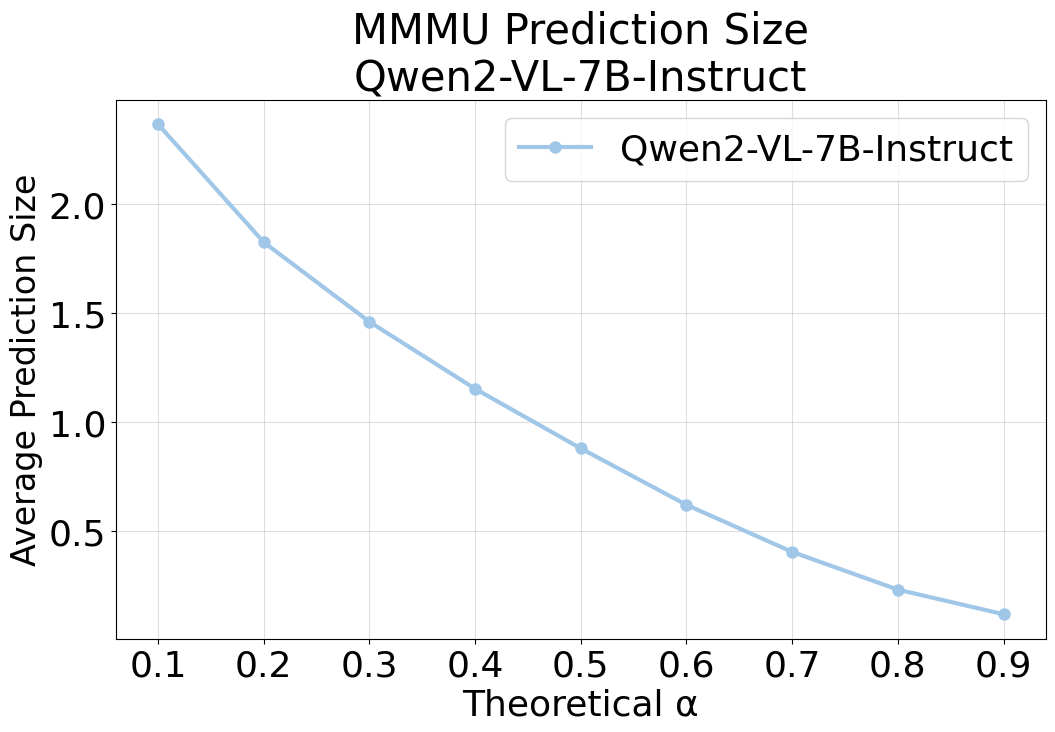}
} &
\subfloat[]{
    \includegraphics[width=5\textwidth]{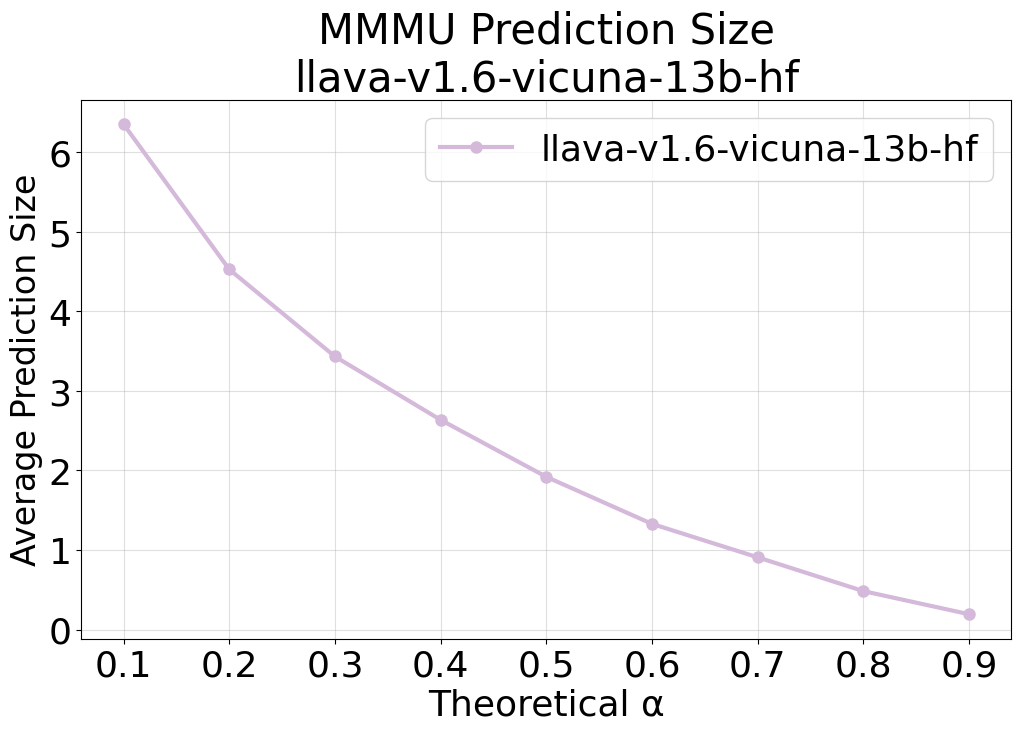}
} \\
\end{tabular}
\end{adjustbox}
\caption{Prediction Set Size in ScienceQA and MMMU Benchmark.Prediction Set Size in ScienceQA and MMMU Benchmark. We calculate the Prediction Set Size and plot it as a solid line in the figure. The x-axis represents the $\alpha$ values, while the y-axis corresponds to the Prediction Set Size.
 }
\label{fig:all}

\end{figure}

\subsection{Prediction Set Size}
\textbf{Non-Conformity Scores and Prediction Set Dynamics.} The computation of non-conformity scores, as defined in Equation (1), establishes a direct relationship between the user-specified error tolerance $\alpha$ and the granularity of prediction sets. Lower $\alpha$ values correspond to higher permissible error rates, which necessitates broader prediction sets to encapsulate a wider range of candidate options. This design ensures that the model's predictions remain statistically valid even under relaxed error constraints. Conversely, higher $\alpha$ values impose stricter error control, compelling the model to generate compact prediction sets that exclude ambiguous or low-confidence options. The interplay between $\alpha$ and set size reflects a fundamental trade-off: stringent error thresholds reduce prediction uncertainty but may discard potentially valid candidates, whereas lenient thresholds preserve inclusivity at the cost of higher empirical error rates.

\textbf{Fixed Split-Ratio (0.5) Analysis.} Under the fixed calibration-to-test split ratio of 1:1, we systematically evaluate prediction set sizes across both datasets and all eight LVLMs (Fig.2). For the ScienceQA dataset, the observed trends align with theoretical expectations: prediction set sizes exhibit a monotonic decrease as $\alpha$ increases, driven by progressively stricter error control. Notably, the rate of size reduction diminishes at lower $\alpha$ thresholds, suggesting an asymptotic stabilization of set sizes under extreme error tolerance. A striking deviation occurs with the Qwen2-VL-7B-Instruct model, where prediction set sizes plateau beyond $\alpha=0.6$. This saturation indicates that the model's predictions approach near-perfect alignment with ground truths, leaving minimal room for further error reduction. Furthermore, the symmetric inverse relationship between empirical error rates (Fig.1) and prediction set sizes (Fig.2) emerges from the marginal coverage constraints in Equation (5). Small perturbations in empirical error rates induce proportional adjustments to set sizes, ensuring adherence to the prescribed statistical guarantees.

\textbf{MMMU Dataset Observations.} While the MMMU dataset broadly mirrors the trends observed in ScienceQA, the InternVL2-1B model exhibits anomalous behavior at $\alpha=0.1$, producing disproportionately large prediction sets. This phenomenon traces to the quantile estimation process detailed in Section 3.2.2: when computing the $0.9$-quantile threshold, a significant concentration of non-conformity scores near 1 artificially inflates the inclusion criterion. Consequently, nearly all candidate options satisfy the relaxed threshold, leading to bloated prediction sets. In contrast, smaller prediction sets arise when non-conformity scores cluster near 0, reflecting high model confidence in its predictions. This dichotomy underscores the sensitivity of conformal prediction methods to the distribution of non-conformity scores, particularly under extreme $\alpha$ values. The observed anomalies highlight the importance of robust score calibration and threshold selection in maintaining practical utility while preserving theoretical guarantees.

\newpage
\section{Conclusion}
We propose a statistical reliability framework based on Split Conformal Prediction to address hallucinations in large vision–language models  on visual question answering tasks. By employing dynamic threshold calibration and cross‑modal consistency verification, we split the data into calibration and test sets, quantify output uncertainty with a nonconformity score, and construct prediction sets from calibration‑set quantiles. At a user‑specified risk level $\alpha$, our method strictly controls the marginal coverage of true answers. Experiments on multiple multimodal benchmarks across diverse LVLM architectures show that SCP meets the theoretical statistical guarantee for all $\alpha$ values, and that prediction‑set size adjusts inversely with $\alpha$, effectively filtering out low‑confidence outputs. Requiring no prior distributional assumptions or model retraining, our model‑agnostic and computationally efficient framework offers solid theoretical and practical support for reliable multimodal evaluation in safety‑critical scenarios.

\bibliographystyle{unsrtnat}
\bibliography{references}  






\end{document}